\documentclass[10pt,twocolumn,letterpaper]{article}

\usepackage[pagenumbers]{cvpr}

\newcommand{\red}[1]{{\color{red}#1}}

\usepackage{overpic}
\usepackage{pifont}
\usepackage[dvipsnames]{xcolor}
\usepackage{multirow}
\newcommand{\thRows}[1]{\multirow{3}*{#1}}

\newcommand{\tRows}[1]{\multirow{2}*{#1}}

\newcommand{\figref}[1]{Fig.~\ref{#1}}
\newcommand{\tabref}[1]{Tab.~\ref{#1}}
\newcommand{\secref}[1]{Section~\ref{#1}}
\newcommand{\cmark}{\ding{51}}
\newcommand{\xmark}{\ding{55}}

\newcommand{\ourMthd}{MTE}
\newcommand{\myPara}[1]{\noindent\textbf{#1}}
\newcommand{\gray}[1]{{\textcolor{gray}{#1}}}

\definecolor{cvprblue}{rgb}{0.21,0.49,0.74}
\usepackage[pagebackref,breaklinks,colorlinks,allcolors=cvprblue]{hyperref}

\author{Zhong-Yu Li$^1$, Yu-Song Hu$^1$, Bo-Wen Yin$^1$, Ming-Ming Cheng$^1$ \\
$^1$VCIP, School of Computer Science, Nankai University \\
}

\title{Multi-Token Enhancing for Vision Representation Learning}

\begin{document}

\maketitle

\begin{abstract} 
    Vision representation learning, especially self-supervised learning, 
    is pivotal for various vision applications. 
    Ensemble learning has also succeeded in enhancing the performance and robustness of the vision models. 
    However, traditional ensemble strategies are impractical for representation learning, 
    especially self-supervised representation learning that 
    requires large-scale datasets and long schedules. 
    This is because they require $k$ times more training and inference computation costs for an ensemble of $k$ models. 
    Differently, 
    we introduce \textbf{M}ulti-\textbf{T}oken \textbf{E}nhancing~(\textbf{\ourMthd})  
    that extracts multiple auxiliary tokens simultaneously 
    from a single model to enhance representation learning, 
    while incurring minimal additional training costs and no additional inference costs. 
    These auxiliary tokens, 
    including auxiliary CLS tokens and adaptively pooled tokens, 
    capture complementary information due to their differences. 
    Meanwhile, to address the increase in inference costs, we distill the knowledge acquired by the auxiliary tokens into a global token during pre-training. 
    Consequently, we can discard the auxiliary tokens during inference without incurring additional costs. 
    Our \ourMthd~is compatible with various self-supervised loss functions and architectures, 
    consistently improving performances across different downstream tasks. 
    Our source code will be made publicly available.
\end{abstract}

\section{Introduction}\label{sec:introduction}

Vision representation learning plays a pivotal role in the field of computer vision. 
Through pre-training on large-scale datasets, the representations have strong performance and robustness across a wide range of vision applications~\cite{caron2021emerging,zhou2021ibot}, \eg image recognition, semantic segmentation, and object detection. 
To further enhance the representation quality, we explore the model ensemble in the field of representation learning 
because that model ensemble has already achieved great success in various vision tasks~\cite{Simpson_2022_CVPR,izmailov2018averaging,GANAIE2022105151}. 
However, traditional ensemble methods~\cite{58871,dietterich2000ensemble} necessitate $k$ times more training or inference computation for an ensemble of $k$ different models. 
Such a substantial resource requirement is often impractical for representation learning, especially self-supervised representation learning that involves training models with large-scale datasets and long schedules. 
In this study, we propose a method wherein a single model is trained to extract multiple different vision representations, thereby enabling the ensemble from a single model 
with minimal additional training costs and no extra inference costs. 

\begin{figure}[t]
  \centering
  \begin{overpic}[width=0.95\columnwidth]{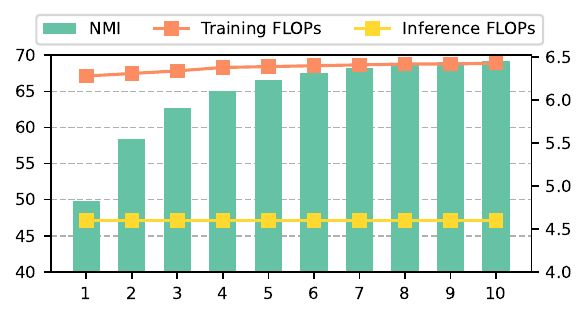}
    \put(21.8, -1){The number of auxiliary tokens}
    \put(-1.5, 18.5){\rotatebox{90}{NMI~(\%)}}
    \put(99, 17){\rotatebox{90}{FLOPS (G)}}
  \end{overpic}
  \caption{
    Increasing the number of the proposed auxiliary tokens leads to greater improvements 
    without incurring additional inference costs. 
    We combine varying numbers of the auxiliary tokens for clustering~(using the prototype layer 
    proposed in \cite{caron2021emerging}) and assess the performance using the normalized mutual information (NMI) between the generated pseudo labels and true labels. 
    As the number of auxiliary tokens increases, they complement each other, resulting in enhanced performances. 
    During inference, the auxiliary tokens are removed without any additional inference costs.
  }
  \label{fig:introduction}
\end{figure}

Specifically, we propose Multi-Token Enhancing~(\ourMthd), which utilizes multiple auxiliary tokens for ensemble learning to learn a powerful representation encoder. 
These tokens are then removed to avoid additional costs during inference. 
Firstly, we devise two types of auxiliary tokens capable of encoding image information. 
These include auxiliary CLS tokens, which consist of multiple trainable CLS tokens~\cite{dosovitskiy2020vit} with different initializations, 
and adaptively pooled tokens with unique pooling weights.
Notably, we discover differences between these auxiliary tokens, enabling them to complement each other in learning high-quality semantic representations, as shown in \figref{fig:introduction} where more auxiliary tokens achieve higher improvements. 
During pre-training, these auxiliary tokens only incur minimal computation costs, \eg an increase from 6.1G to 6.4G floating-point operations per second in the training forward of a $224 \times 224$ image. 

Meanwhile, to circumvent the additional inference costs brought by the auxiliary tokens, we propose an online distillation strategy, which transfers the knowledge acquired by the auxiliary tokens into a global token during pre-training. 
After pre-training, the global token effectively acquires the majority of the knowledge 
held by the auxiliary tokens. 
Consequently, we can eliminate the auxiliary tokens to save inference costs without compromising performance. 
Surprisingly, our experiments reveal that the online distillation not only benefits the global token but also enhances the auxiliary tokens. 
This occurs due to their shared encoder, leading to coupled optimization of the global token and auxiliary tokens. 
Moreover, stronger auxiliary tokens, serving as the distillation teacher, further enhance the global token. 
As a result, the distillation builds a reciprocal optimization dynamic where the global token and auxiliary tokens promote each other. 

\begin{figure}[t]
  \centering
  \begin{overpic}[width=0.95\columnwidth]{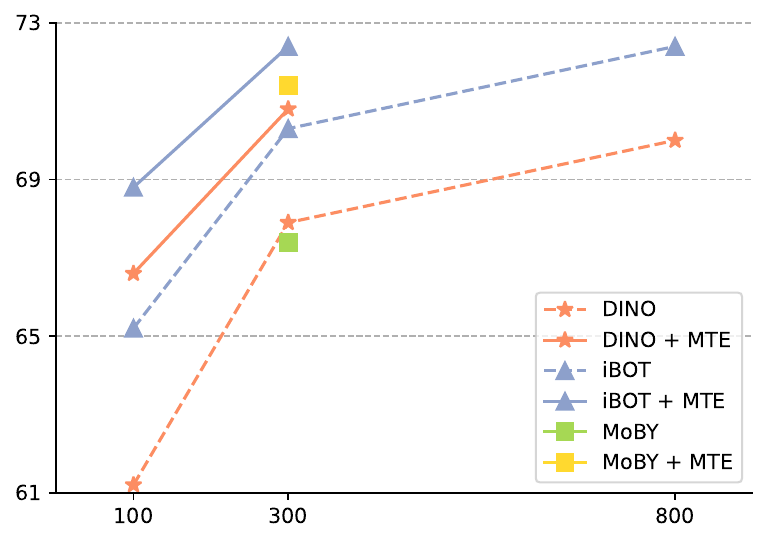}
    \put(35, -2){Pre-training Epochs}
    \put(-3, 33){\rotatebox{90}{Top-1}}
  \end{overpic}
  \caption{
    $k$-NN Top-1 accuracies when cooperating \ourMthd~with different methods, 
    including MoBY~\cite{xie2021moby}, 
    DINO~\cite{caron2021emerging}, 
    and iBOT~\cite{zhou2021ibot}.
  }
  \label{fig:epoch_performance}
\end{figure}

Our proposed method can be implemented using various loss functions for self-supervised and supervised learning.
Across various downstream tasks, including image classification, semantic segmentation, 
and instance segmentation, \ourMthd~consistently boosts performances. 
For example, \figref{fig:epoch_performance} shows that our \ourMthd~performs better with fewer training epochs. 

Our major contributions are summarized as follows:
\begin{itemize}
  \item We propose Multi-Token Enhancing~(\ourMthd), which leverages multiple auxiliary tokens to capture diverse and complementary image representations, thus enhancing representation learning. Our \ourMthd~is orthogonal with various loss functions and architectures, consistently boosting performance across different downstream tasks.
  \item We propose an online distillation strategy that allows the pre-trained models   to discard the auxiliary tokens during inference without compromising performance. 
\end{itemize}

\section{Related Works}
\label{sec:related_works}

\subsection{Self-supervised Learning}
The self-supervised learning methods can be categorized into two 
primary categories, 
\ie masked image modeling~\cite{Woo2023ConvNeXtV2,Xie_2022_CVPR,cwei2021,Chen_2023_CVPR,Feng_2023_CVPR} and 
instance discrimination~\cite{oord2018representation,li2023sere,li2023enhancing,caron2021emerging}. 
Among them, 
instance discrimination 
generates different views of an image via random image augmentations 
and pulls their representations together. 
Based on this basic framework, 
various forms of loss functions are proposed, 
such as contrastive loss 
\cite{DecoupledContrastive,Xie_2021_ICCV,Zhu_2023_ICCV,Song_2023_ICCV}, 
similarity loss
\cite{DynamicsContrastive,Chen_2021_CVPR,ermolov2021whitening,Nakamura_2023_ICCV}, 
and self-clustering loss~\cite{Zhan_2020_CVPR,yangCVPR2016joint,asano2020self,song2023multimode}. 
In this work, 
we further enhance these self-supervised methods 
through the ensemble of multiple auxiliary tokens, 
which are extracted from a single model with minimal training costs. 
This differs from existing methods~\cite{Chen_2021_ICCV,He_2020_CVPR,wang2020DenseCL,Xie_2021_CVPR,Tao_2023_CVPR} 
that leverage only a single token, 
\ie a CLS token~\cite{zhou2021ibot,Chen_2021_ICCV,mugs2022SSL} 
or a pooled token~\cite{He_2020_CVPR,chen2020simple}, 
to learn global representations.

\def \adpP {$\mathcal{P}$}

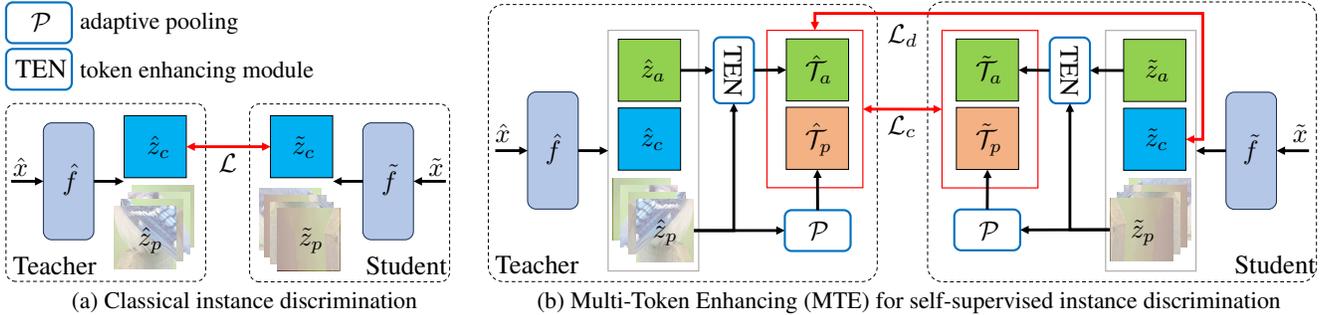
\begin{figure*}[t]
  \centering
  \begin{overpic}[height=.215\linewidth]{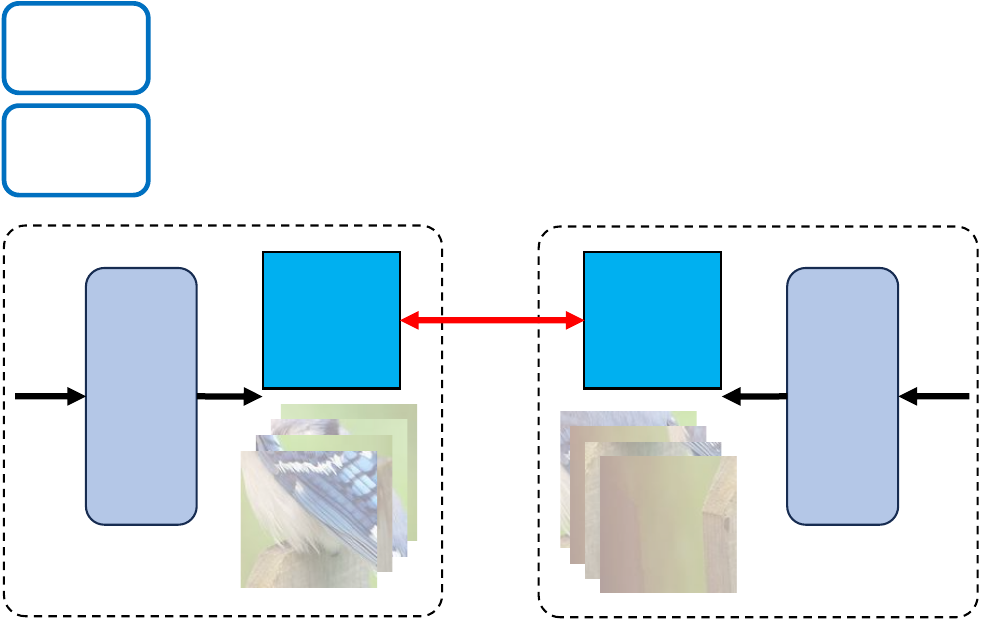}
    \put(17,57){\small adaptive pooling}
    \put(17,46){\small token enhancing module}
    \put(6,56.5){\adpP}
    \put(2.5,46){TEN}
    \put(2, 2){Teacher}
    \put(81,2){Student} 
    \put(13,21){$\hat{f}$}
    \put(84.5,21){$\tilde{f}$}
    \put(2, 24){$\hat{x}$}
    \put(95,24){$\tilde{x}$}
    \put(32,29){$\hat{z}_c$}
    \put(64.5,29){$\tilde{z}_c$} 
    \put(30,8.5){$\hat{z}_p$} 
    \put(66,8.5){$\tilde{z}_p$} 
    \put(48,25){$\mathcal{L}$}
  \end{overpic} \hfill
  \begin{overpic}[height=.215\linewidth]{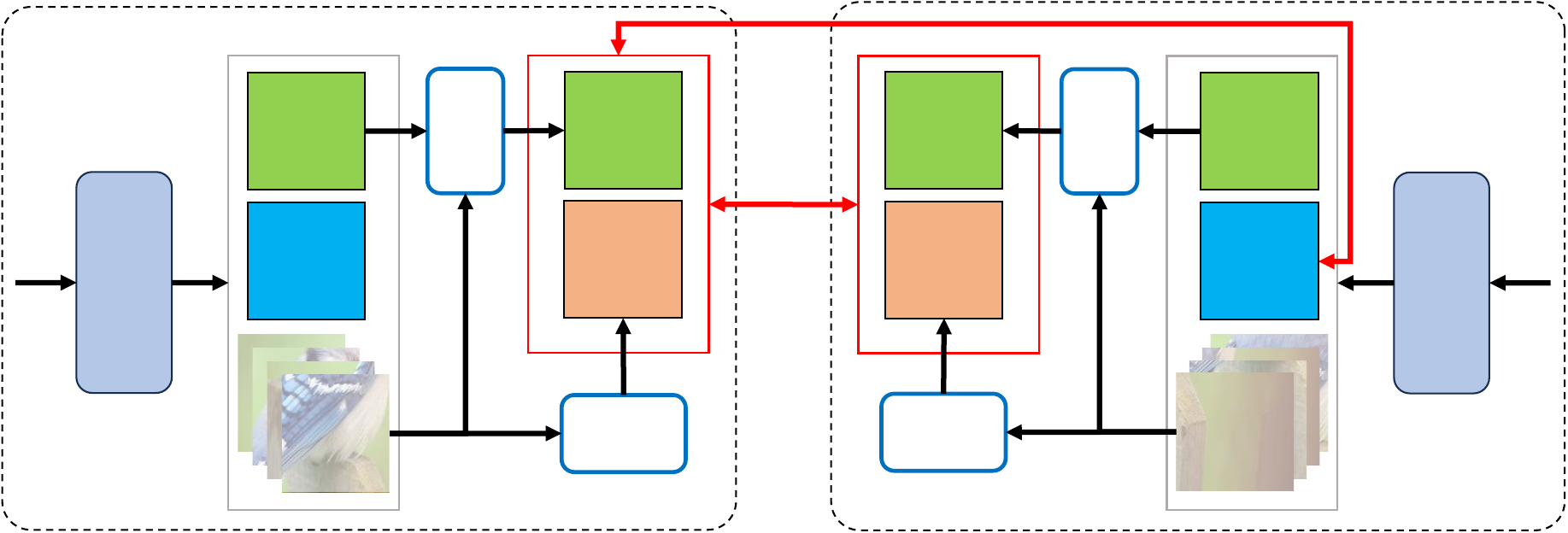}
    \put(-50,-3){\small (a) Classical instance discrimination}
    \put(6,-3){\small (b) Multi-Token Enhancing~(\ourMthd) for self-supervised 
    instance discrimination}
    \put(1,1.2){Teacher}
    \put(90,1.2){Student} 
    \put(7,15){$\hat{f}$} 
    \put(91,15){$\tilde{f}$}
    \put(1,17){$\hat{x}$}
    \put(97,17){$\tilde{x}$}
    \put(18.5,16.5){$\hat{z}_c$}
    \put(79.2,16.5){$\tilde{z}_c$}
    \put(20,5.5){$\hat{z}_p$}
    \put(77.5,5.5){$\tilde{z}_p$}
    \put(28.5,28.6){\rotatebox{270}{TEN}}
    \put(69,28.6){\rotatebox{270}{TEN}}
    \put(38.7,5.3){\adpP}
    \put(59.2,5.3){\adpP}
    \put(18.5,24.8){$\hat{z}_a$}
    \put(79.2,24.8){$\tilde{z}_a$}
    \put(38.3,24.5){$\hat{\mathcal{T}}_a$}
    \put(58.8,24.5){$\tilde{\mathcal{T}}_a$}
    \put(38.3,16.3){$\hat{\mathcal{T}}_p$}
    \put(58.8,16.3){$\tilde{\mathcal{T}}_p$}
    \put(48.2,29){$\mathcal{L}_{d}$}
    \put(48.2,18){$\mathcal{L}_{c}$}
  \end{overpic}\\ \vspace{8pt}
  \caption{For effective training, our \ourMthd~employs additional 
    auxiliary parts,
    which will be discarded during inference and fine-tuning.
  }\label{fig:framework}
  \vspace{-15pt}
\end{figure*}

\subsection{Vision Architectures}
In recent years, 
various architectures have been proposed, 
with convolution neural networks~\cite{he2016deep,liu2022convnet} and vision transformers~\cite{dosovitskiy2020vit,wang2021pyramid,liu2021Swin} 
as two main categories. 
To enhance the generalization ability on downstream tasks, 
these architectures are usually pre-trained on large-scale datasets, 
where they substitute a CLS token~\cite{zhou2021ibot,Chen_2021_ICCV,mugs2022SSL} 
or a pooled token~\cite{He_2020_CVPR,chen2020simple} 
into the loss functions of self-supervised or supervised loss functions 
to learn the ability to encode image information. 
In this work, 
we further enhance the capacity of these architectures from the perspective of the model ensemble 
without training multiple models. 
Instead of leveraging only a single token, we make a single model output a series of auxiliary tokens, 
which have differences and thus are complementary to encode representations of high quality. 
Our method is compatible with different architectures, including convolution and transformer models. 
After training, these auxiliary tokens are removed without changing the model structure and increasing inference costs.

\subsection{Model Ensemble}
Model ensemble\cite{58871,dietterich2000ensemble} has been explored in deep learning and 
improved model performance and robustness across various tasks. 
Traditional methods train multiple models independently and 
integrate the predictions of all models during inference. 
However, the ensemble of multiple models brings huge computation costs in training and inference and 
thus is not feasible 
for representation learning, 
especially self-supervised representation learning that 
usually trains the model with large-scale datasets and long schedules. 
To solve this problem, 
some works~\cite{izmailov2018averaging} propose to average the parameters of multiple models 
to approximate model ensemble. 
In self-supervised~\cite{He_2020_CVPR,caron2021emerging} and supervised learning~\cite{liu2021Swin,liu2022convnet}, 
the exponential moving average, which ensemble models from different training iterations, 
has already been widely adopted. 
Additionally, 
some works propose the ensemble with shared parameters~\cite{havasi2021training,lee2015m,Wen2020BatchEnsemble} 
to optimize training or inference efficiency. 
Some regularization methods, \eg dropout~\cite{Dropout} and drop-path~\cite{larsson2017fractalnet}, 
can also be viewed as the parameter-shared ensemble of sub-networks sampled during training 
and have already been utilized in representation learning to train large models. 
In this study, 
we adopt a single model to extract different auxiliary tokens that also share the encoder parameters, 
thus with minimal additional training costs. 
Meanwhile, 
we remove the auxiliary tokens during inference without sacrificing the performance. 
Thus, the model has no additional inference costs. 

\subsection{Self-supervised Knowledge Distillation}
In this work, 
we propose an online distillation that 
transfers the knowledge held by multiple auxiliary tokens 
into a global token, 
enabling us to remove the auxiliary tokens during inference. 
This is related to the knowledge distillation~\cite{hinton2015distilling} field, 
which transfers knowledge from a large teacher 
to a small student and has already achieved 
tremendous progress. 
Recently, 
aiming to solve the problem that 
self-supervised learning does not work well on small models, 
some self-supervised methods~\cite{fang2021seed,navaneet2021simreg,xu2022bag} 
utilize knowledge distillation to transfer pre-trained fixed models 
to small models.  
Some self-distillation-based methods~\cite{byol,caron2021emerging} 
dynamically update the teacher according to the student. 
In this work, 
we use auxiliary tokens as 
the teacher and the global token as the student, 
and we also dynamically update the auxiliary tokens along with the optimization of the global token. 
Thus, 
we build a reciprocal optimization between the auxiliary tokens and the global token.
Meanwhile, 
this way could be seamlessly integrated into 
existing self-supervised frameworks
based on siamese networks~\cite{He_2020_CVPR,caron2021emerging}, 
where we can use two siamese networks 
as teacher and student. 

\begin{table}[t]
  \centering
  \setlength{\tabcolsep}{1.8mm}
  \caption{Table of symbols, their dimensions, and meaning.}
  \begin{tabular}{ccc} \hline
    Symbol & Dim. & Meaning \\ \hline
    $D,N,M,K$ & scalar & dimension/number of tokens  \\
    $\mathcal{L}, \mathcal{L}_{c}, \mathcal{L}_{d}$ & & 
    loss functions in Equ. (\ref{eq:abtract_loss}, 
    \ref{eq:loss_instance_discrimination}, \ref{eq:loss_distilltion})\\
    $\hat{z}_c$ & $1\times D$ & (global) CLS token \\
    $\hat{z}_p$ & $N\times D$ & patch tokens \\
    $\hat{z}_a$ & $M\times D$ & auxiliary CLS tokens \\
    $\hat{\mathcal{T}_a}$ & $M\times D$ & enhanced auxiliary CLS tokens\\
    $\hat{\mathcal{T}_p}$ & $K\times D$ & adaptively pooled tokens \\ \hline
  \end{tabular}
  \label{tab:Symbols}
  \vspace{-10pt}
\end{table}

\section{Method}
\label{sec:method}

In this section, 
we focus on self-supervised learning 
because of its strong capacity to learn representations. 
In \secref{sec:preliminaries}, 
we first introduce the classical self-supervised learning, as shown in \figref{fig:framework} (a). 
Different from the classical method, 
as shown in \figref{fig:framework} (b), 
our proposed \ourMthd~introduces 
two types of auxiliary tokens~(defined in \secref{sec:multi_token_define}) 
into self-supervised learning in \secref{sec:multi_token_ssl}, 
enabling the models to capture rich image semantics. 
To avoid increasing inference costs, 
we further design an online distillation strategy in \secref{sec:online_distillation}, 
which distills the knowledge of the auxiliary tokens into a global token, 
enabling us to discard the auxiliary tokens for effective inference. 
In \secref{sec:adapt_to_supervised}, 
we also adapt the proposed method to supervised learning. 
Additionally, 
we analyze some factors that make the proposed method effective 
in \secref{sec:analysis}. 

\subsection{Preliminaries}
\label{sec:preliminaries}

In this work, 
we take the vision transformer as an example to describe our proposed \ourMthd. 
\secref{sec:experiments} 
further shows that the \ourMthd~is compatible with other architectures. 

\myPara{Vision transformer.} 
Given an input image $x$ as the input, 
a plain vision transformer~\cite{dosovitskiy2020vit} $f$ 
outputs $[z_c, z_p] = f(x)$, 
where $z_{c} \in \mathbb{R}^{1 \times D}$ is the CLS token, 
$z_{p} \in \mathbb{R}^{N \times D}$ are the patch tokens, 
and $D$ means the number of channels. 
The CLS token is usually trained to encode global context among these tokens.

\myPara{Self-supervised learning.} 
Given an image $x$, 
self-supervised learning first 
uses random data augmentations to 
generate distinct views~\cite{chen2020simple,He_2020_CVPR} 
and pull them together in the representation space. 
For clarity, 
we consider two views, \ie $\hat{x}$ and $\tilde{x}$, 
whose representations are extracted by the teacher network $\hat{f}$ 
and student network $\tilde{f}$ as 
$[\hat{z}_{c}, \hat{z}_{p}]=\hat{f}(\hat{x})$ and $[\tilde{z}_{c}, \tilde{z}_{p}]=\tilde{f}(\tilde{x})$, 
respectively. 
Most methods~\cite{He_2020_CVPR,caron2021emerging} further process the CLS tokens by projection heads, 
\ie $\hat{h}_{c}=\hat{p}(\hat{z}_{c})$ and $\tilde{h}_{c}=\tilde{p}(\tilde{z}_{c})$. 
Subsequently, they are substituted into different forms of loss functions, 
\eg InfoNCE loss~\cite{chen2020simple, Chen_2021_ICCV, oord2018representation}, 
clustering loss~\cite{caron2020unsupervised,caron2021emerging}, 
similarity loss~\cite{byol,Chen_2021_CVPR}. 
To simplify, 
we abstract these loss functions as follows:
\begin{equation}\label{eq:abtract_loss}
    {\mathcal{L}}(\hat{h}_{c}, \tilde{h}_{c}).    
\end{equation}
Some methods~\cite{Chen_2021_CVPR} also attach a predictor head 
to the student network, 
and we omit it for conciseness. 

\subsection{Auxiliary Tokens}
\label{sec:multi_token_define}

Existing architectures often adopt 
a CLS token~\cite{dosovitskiy2020vit,Wu_2021_ICCV,convit} or 
a pooled token~\cite{wu2022p2t,he2016deep,liu2022convnet} 
to encode global context. 
We further extend these two types to form multiple auxiliary tokens, 
which can be complementary to enhance representations. 
Specifically, the CLS token is extended to multiple with different random initializations. 
The pooled token is also extended 
to multiple with distinct adaptive weights. 
In \secref{sec:analysis_difference}, 
we will further analyze the complementarity across different auxiliary tokens. 

\myPara{Auxiliary CLS token.} 
Apart from the original CLS token~\cite{dosovitskiy2020vit}, 
we make the transformers output $M$ extra CLS tokens 
$z_a \in \mathbb{R}^{M \times D}$, 
referred to as auxiliary CLS tokens. 
Specifically, 
rather than concatenating only one trainable token with 
the patch tokens in~\cite{dosovitskiy2020vit}, 
the other $M$ trainable tokens that are differently initialized 
are also concatenated with the patch tokens. 
Then, these trainable tokens are fed into stacked transformer blocks 
to capture the global context. 
In summary, the transformer outputs 
$[z_c, z_a, z_p] = f(x)$.

\myPara{Token Enhancing module.} 
\secref{sec:online_distillation} 
will demonstrate that the auxiliary tokens will be distilled 
to another global token and discarded during inference. 
To provide a more powerful distillation teacher, 
we enhance the auxiliary CLS tokens with a Token Enhancing~(TEN) module 
during pre-training. 
Notably, this module will be discarded after pre-training, 
thus not introducing additional parameters during inference. 
Specifically, the TEN module enhances 
auxiliary CLS tokens 
through the attention mechanism. 
Given the auxiliary CLS tokens $z_{a} \in \mathbb{R}^{M \times D}$ 
as the queries and patch tokens $z_{p} \in \mathbb{R}^{N \times D}$ 
as the keys and values, 
they are fed into a multi-head cross-attention layer. 
Then, we use MLP to process each auxiliary CLS token and 
obtain the enhanced auxiliary CLS tokens 
$\mathcal{T}_a={\rm TEN}(z_a, z_p)$, 
which can capture more powerful semantics than the raw ones.

\myPara{Adaptively pooled token.} 
Apart from the CLS token, pooling from patch tokens 
$z_p \in \mathbb{R}^{N \times D}$ can also represent the image semantics. 
However, simple average pooling mixes the features of different objects 
in the foreground and background, 
causing noise and the loss of 
the discriminative object information~\cite{Shu_2023_CVPR}. 
More importantly, 
global pooling can only produce one token, 
limiting the ability to extract multiple complementary tokens. 
Thus, we utilize adaptive weights to pool discriminative regions, 
where the adaptive weights are obtained through 
the large kernel convolution~\cite{hou2022conv2former,guo2022visual}. 
Meanwhile, to acquire different semantic information, 
$K$ different weights are generated by $K$ convolutions of 
different random initializations.
In summary, given the patch tokens $z_{p} \in \mathbb{R}^{N \times D}$, 
we define an adaptive pooling function \adpP~that generates 
$K$ different weights 
$\{W^i \in \mathbb{R}^{N \times D} \mid i \in [1, K] \}$ and 
calculate adaptively pooled tokens 
$\mathcal{T}_p \in \mathbb{R}^{K \times D}$ 
as follows:
\begin{equation}
  \mathcal{T}_p^i = \mathcal{P}(W^{i}_{}, z_p) = {\rm Average}(W^{i}_{} \odot z_{p}), 
\end{equation}
where $\odot$ means element-wise multiplication and 
${\rm Average}$ is the average pooling function. 

\subsection{Multi-Token Self-supervised Learning}
\label{sec:multi_token_ssl}

This section describes how to use the auxiliary tokens for self-supervised learning. 
Given multiple auxiliary tokens, 
we feed them into the projection head following existing methods~\cite{caron2021emerging,He_2020_CVPR} 
and then fuse them to 
acquire a powerful token that 
encodes comprehensive and diverse 
semantics. 
Meanwhile, considering that 
a customized independent head 
can better handle the unique information contained in each auxiliary token, 
we use an independent projection head to process each auxiliary token as follows: 
\begin{equation}
    h_{t} = \frac{\sum_{i=1}^{M}p^i(\mathcal{T}_a^i) + \sum_{i=1}^{K}p^{M+i}(\mathcal{T}_p^i)}{M+K},
    \label{eq:fuse_with_shared_projection}
\end{equation}
where $\{p^i \mid i \in [1, K+M]\}$ 
are $K+M$ independent projection heads. 
Then, we can substitute $h_{t}$ into \eqref{eq:abtract_loss} 
and obtain the loss function as follows: 
\begin{equation}
    {\mathcal{L}_{c}} = {\mathcal{L}}(\hat{h}_{t}, \tilde{h}_{t}). 
    \label{eq:loss_instance_discrimination}
\end{equation}

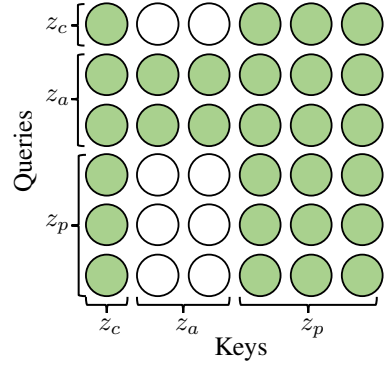
\begin{figure}[t]
  \centering
  \begin{overpic}[width=0.5\linewidth]{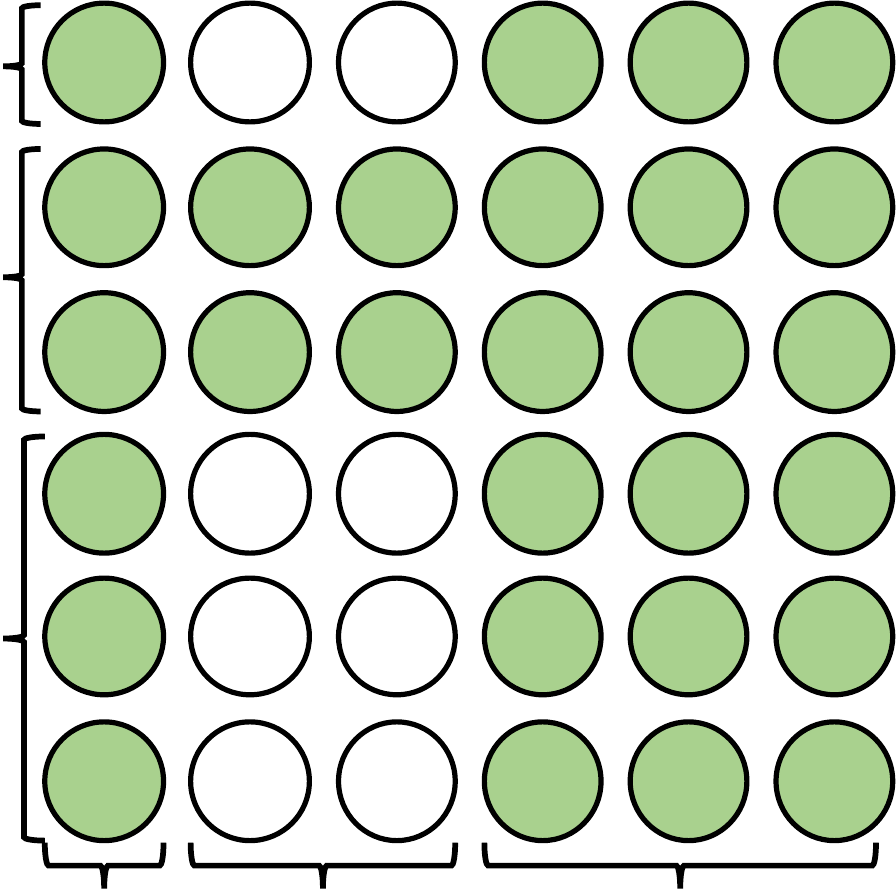}
    \put(46, -13){Keys}
    \put(-18, 39){\rotatebox{90}{Queries}}
    \put(9, -5){$z_c$}
    \put(33.5, -5){$z_a$}
    \put(73.5, -5){$z_p$}
    \put(-8, 91){$z_c$}
    \put(-8, 68){$z_a$}
    \put(-8, 28){$z_p$}
  \end{overpic}
  \vspace{10pt}
  \caption{
    The attention mask in the self-attention layers. 
    $z_c$, $z_a$, and $z_p$ represent the CLS token, auxiliary CLS tokens, and patch tokens, respectively. 
    White circles mean that the corresponding query is not allowed to attend to the corresponding key. 
  }
  \label{fig:attention_mask}
  \vspace{-10pt}
\end{figure}

\subsection{Online Distillation}
\label{sec:online_distillation}

Compared to the single CLS token used by existing methods
\cite{He_2020_CVPR,caron2021emerging}, 
the auxiliary tokens can learn more powerful representations. 
However, they also introduce an extra computational budget. 
For fast inference, 
we propose distilling the knowledge 
held by auxiliary tokens into another global token. 
This distillation operation enables us
to remove those auxiliary tokens 
and use the distilled global token for inference. 

\myPara{Distillation into the global token.} 
When using the plain vision transformer~\cite{dosovitskiy2020vit}, 
we use the original CLS token $z_c$ as the global token. 
To calculate the loss function, 
we also project $z_c$ into $h_c=p_c(z_c)$ by the projection head $p_c$. 
Then, we substitute $\tilde{h}_c$ from the student network and 
$\hat{h}_t$ from the teacher network into the loss function as follows:
\begin{equation}\label{eq:loss_distilltion}
  \mathcal{L}_{d} = {\mathcal{L}}(\hat{h}_t, \tilde{h}_c). 
\end{equation}
This loss will encourage the global token to acquire the knowledge 
held by the auxiliary tokens. 
In particular, this distillation is performed online. 
Therefore, the overall loss function for pre-training 
is defined as follows:
\begin{equation} \label{eq:overall_loss}
  \mathcal{L}={\mathcal{L}}(\hat{h}_t, \tilde{h}_t) + {\mathcal{L}}(\hat{h}_t, \tilde{h}_c).
\end{equation}

Surprisingly, 
we discover that such distillation 
enhances not only the global token 
but also the auxiliary tokens, 
because 
the optimization of auxiliary tokens accompanies the optimization of the global token. 
In \secref{sec:analysis_distillation}, we will analyze this phenomenon in detail. 

\begin{figure*}[t]
  \centering
  \begin{minipage}[t]{0.48\linewidth}
    \centering
    \begin{overpic}[width=1.0\linewidth]{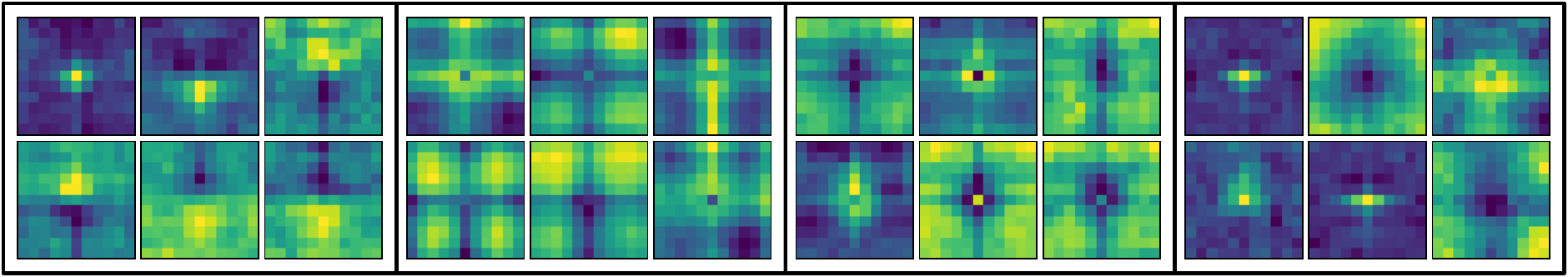}
    \end{overpic}
    \caption{
      The visualization of the kernel weights 
      in the large kernel convolutions used to generate adaptively pooling weights.
      Each black box corresponds to a randomly chosen channel 
      and contains the $11\times 11$ kernel weights 
      of six adaptively pooled tokens.
    }
    \label{fig:kernel_visualization}
  \end{minipage}
  \hspace{5pt}
  \begin{minipage}[t]{0.48\linewidth}
    \centering
    \begin{overpic}[width=1.0\linewidth]{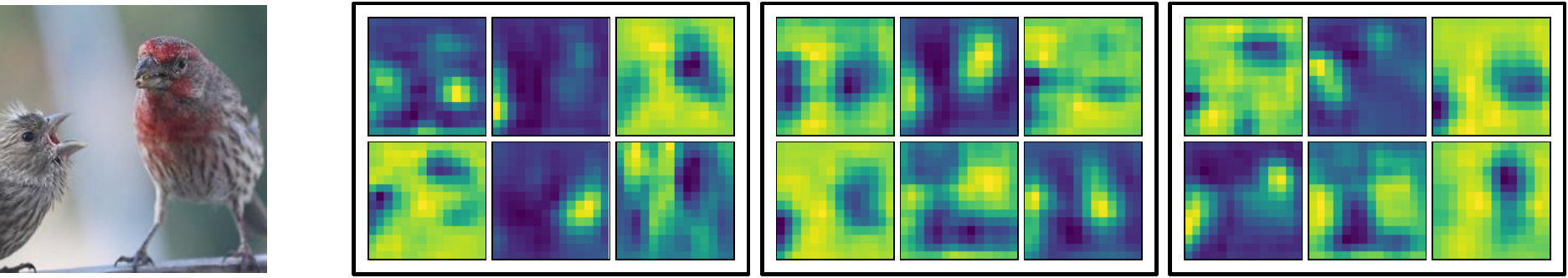}
    \end{overpic}
    \caption{
      The visualization of the adaptive pooling weights in 
      three randomly chosen channels.
      Each black box corresponds to a channel 
      and contains the $14\times 14$ pooling weights 
      of six adaptively pooled tokens.
    }
    \label{fig:activation_visualization}
  \end{minipage}
\end{figure*}

\myPara{Inference of lossless performance.} 
After pre-training, 
we remove the auxiliary tokens to 
avoid extra costs during inference. 
Meanwhile, it should be ensured that 
removing them will not sacrifice performance. 
This condition is naturally satisfied 
for the architectures using global average pooling as the global token, 
because the calculation of patch tokens and the global token
has nothing to do with the auxiliary tokens. 
However, for the plain vision transformer~\cite{dosovitskiy2020vit}, 
we should mask the attention in the multi-head self-attention modules to 
prevent 
the global CLS token and the patch tokens 
from learning to rely on the auxiliary CLS tokens during pre-training, 
as shown in  \figref{fig:attention_mask}. 
With the masking operation, 
the CLS token and patch tokens have consistent behavior when training and inference 
and thus their representations will not be weakened even though the auxiliary CLS tokens are unavailable.

\subsection{Adapting to Supervised Learning}
\label{sec:adapt_to_supervised}

Apart from self-supervised representation learning, 
we also utilize multi-token enhancing 
to enhance supervised representation learning. 
Like self-supervised learning in \secref{sec:multi_token_define}, 
an encoder $f$ output $M$ auxiliary CLS tokens $\mathcal{T}_a \in \mathbb{R}^{M \times D}$ and 
$K$ adaptively pooled tokens $\mathcal{T}_p \in \mathbb{R}^{K \times D}$. 
Like \secref{sec:multi_token_ssl} that uses different projection heads 
for different auxiliary tokens, 
we also process different auxiliary tokens 
with different classifiers as follows:
\begin{equation}
    l_{t} ={\rm Softmax}\left(\frac
    {\sum_{i=1}^{M}(\mathcal{T}_a^i \cdot W_c^i) 
    + \sum_{i=1}^{K}(\mathcal{T}_p^i \cdot W_c^{M+i})
    }{M+K}\right),
    \label{eq:sup_aux_fuse}
\end{equation}
where $\{W_1^i \in \mathbb{R}^{D \times C} \mid i \in [1, M+K] \}$ are the weights of unshared classifiers 
and $C$ is the number of categories. 
Meanwhile, the output regarding the global token is calculated as follows: 
\begin{equation}
    l_c={\rm Softmax}(z_c \cdot W_c).
\end{equation}
Then, 
we supervise the auxiliary tokens with the ground truth and 
distill the auxiliary tokens into the global token. 
The total loss function is shown as follows:
\begin{equation}
{\mathcal{L}}_{sup} = {\rm CE}(l_t, y) + {\rm CE}(l_c, {\rm sg}(l_t)),
\end{equation}
where ${\rm CE}$ is the cross-entropy loss and 
${\rm sg}$ is the gradient detach operation. 
During inference, 
the auxiliary tokens are removed to save computational costs.

\section{Method Analysis}
\label{sec:analysis}

In this section, 
we analyze some factors that make the proposed method effective, 
including the difference and complementarity between auxiliary tokens in \secref{sec:analysis_difference}, 
the reciprocal optimizing process built by the online distillation in \secref{sec:analysis_distillation}, 
and the impacts on patch representations in \secref{sec:analysis_patch}. 
Unless otherwise stated, 
we pre-train models with four auxiliary CLS tokens and six adaptive pooled tokens.

\begin{figure}[t]
  \centering
  \begin{overpic}[width=0.95\linewidth]{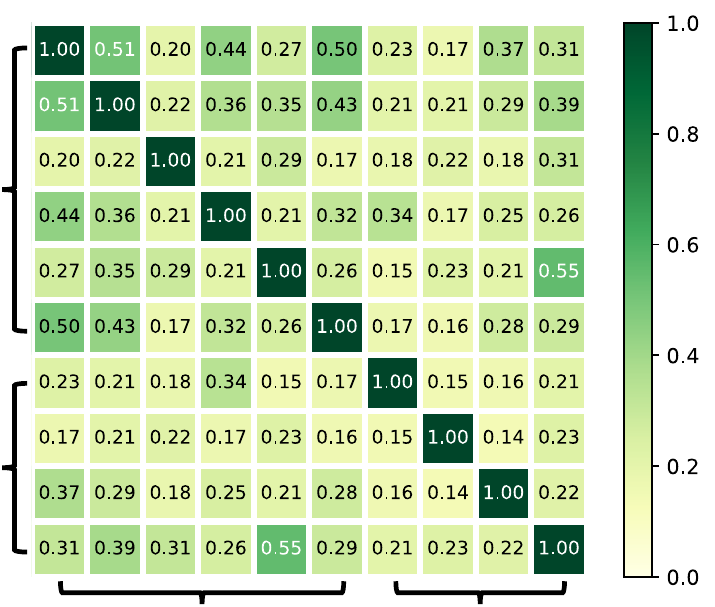}
    \put(26.5, -5){$\mathcal{T}_p$}
    \put(65.5, -5){$\mathcal{T}_a$}
    \put(-5, 57){$\mathcal{T}_p$}
    \put(-5, 18){$\mathcal{T}_a$}
  \end{overpic}
  \vspace{5pt}
  \caption{
    The centered kernel alignment~(CKA)~\cite{kornblith2019similarity} between different pairs of auxiliary tokens 
    when using independent projection heads 
    for different auxiliary tokens. 
    We measure the CKA using the auxiliary tokens output by the projection heads.
    $\mathcal{T}_a$ and $\mathcal{T}_p$ are the auxiliary CLS tokens and adaptively pooled tokens, respectively. 
  }
  \label{fig:cka}
\end{figure}

\begin{figure}[t]
  \centering
  \begin{subfigure}[t]{0.48\linewidth}
    \begin{overpic}[width=1.0\columnwidth]{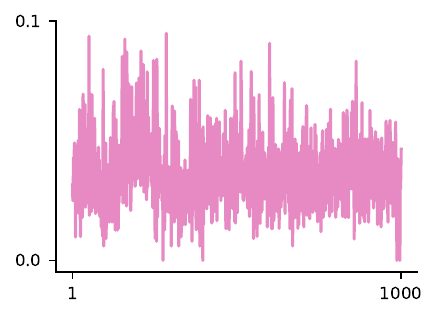}
      \put(35, -4){\small Category IDs}
      \put(-3, 11.0){\rotatebox{90}{\small standard deviation}}
    \end{overpic}
  \end{subfigure}
  \begin{subfigure}[t]{0.48\linewidth}
    \begin{overpic}[width=1.0\columnwidth]{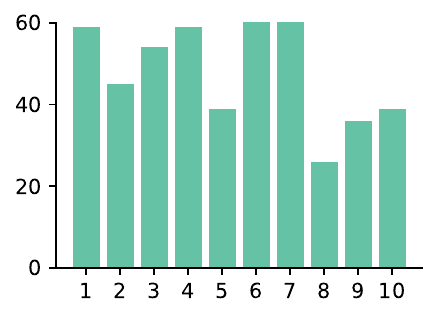}
      \put(-3, 27.5){\rotatebox{90}{\small number}}
      \put(26, -4){\footnotesize Auxiliary Token IDs}
    \end{overpic}
  \end{subfigure}
  \caption{
  (a) For each category, 
  the standard deviation of the accuracies achieved by different auxiliary tokens. 
  (b) 
  The number of categories on which a certain auxiliary token  
  surpasses all other auxiliary tokens. 
  In (b), 
  we only count the 504 categories with only one best-performing token. 
  }
  \label{fig:difference_in_classification}
\end{figure}

\begin{figure}[t]
  \centering
  \begin{overpic}[width=1.0\linewidth]{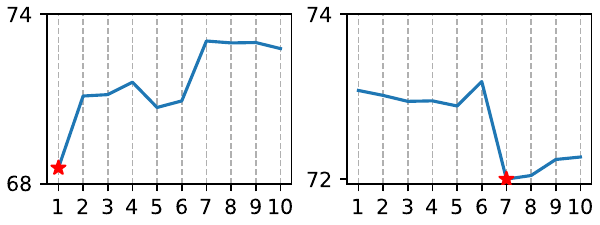}
    \put(12.5, -3){Auxiliary Token IDs}
    \put(62.5, -3){Auxiliary Token IDs}
    \put(1, 13.5){\rotatebox{90}{Top-1~(\%)}}
    \put(52, 13.5){\rotatebox{90}{Top-1~(\%)}}
  \end{overpic}
  \caption{
    The complementarity between auxiliary tokens. 
    We combine the \textbf{first}/\textbf{seventh} auxiliary token with different auxiliary tokens~(represented by ID 1-10 on the horizontal axis) 
    and report the linear probing accuracies in the \textbf{left}/\textbf{right} chart. 
    The red star represents the accuracy without combining different tokens. 
    The first token is an auxiliary CLS token and 
    the seventh token is an adaptively pooled token. 
    The visualizations for more auxiliary tokens are shown in the supplementary material. 
  }
  \label{fig:complementary_linear_probing}
\end{figure}

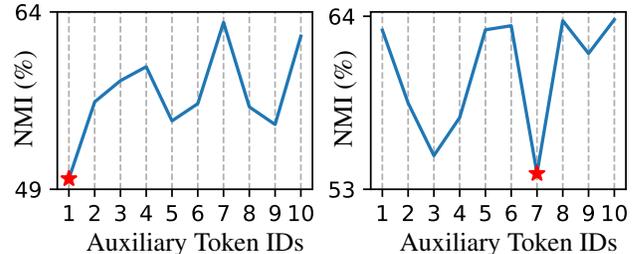
\begin{figure}[t]
  \centering
  \begin{overpic}[width=1.0\linewidth]{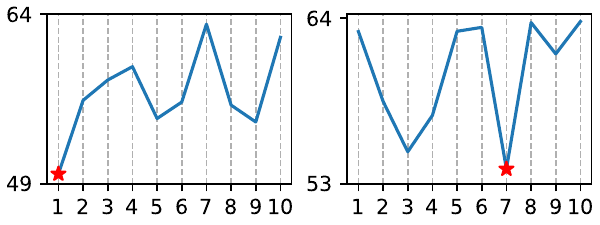}
    \put(12.5, -3){Auxiliary Token IDs}
    \put(62.5, -3){Auxiliary Token IDs}
    \put(1, 13.5){\rotatebox{90}{NMI~(\%)}}
    \put(52, 13.5){\rotatebox{90}{NMI~(\%)}}
  \end{overpic}
  \caption{
    The complementarity between auxiliary tokens after the projection heads. 
    The meaning of the data is consistent with that in \figref{fig:complementary_linear_probing}, 
    except that 
    we feed the combined tokens into the pre-trained prototype layer~\cite{caron2021emerging} to acquire pseudo labels 
    and report the normalized mutual information~(\%) 
    between the pseudo labels and true labels. 
  }
  \label{fig:complementary_nmi}
\end{figure}

\subsection{Complementarity between Auxiliary Tokens}
\label{sec:analysis_difference}
In this section, 
we first show the differences between learned tokens 
from different aspects  
and further show the complementarity.

\myPara{Quantitative difference.} 
To quantify the differences, 
we calculate the centered kernel alignment~(CKA)~\cite{kornblith2019similarity} 
between auxiliary tokens, 
where a lower CKA means more differences. 
As shown in \figref{fig:cka}, 
there are varying degrees of difference between different tokens.

\myPara{Qualitative differences.} 
In \secref{sec:multi_token_define}, 
we use different convolutions to generate adaptive pooling weights. 
\figref{fig:kernel_visualization} shows that 
the convolutions corresponding to different adaptive pooled tokens 
have distinct kernel weights and 
thus output different pooling weights  
in \figref{fig:activation_visualization}. 
We observe that different tokens attend different semantic regions,  
enabling the acquisition of different semantic representations. 

\myPara{Semantic differences.} 
We also discover that 
different tokens 
perform differently in each category, 
as shown in \figref{fig:difference_in_classification} (a). 
Meanwhile, 
different auxiliary tokens 
are expert at recognizing different categories, 
verified by \figref{fig:difference_in_classification} (b), 
which shows the number of categories in which 
each token outperforms all other tokens. 
These results 
imply that different auxiliary tokens can be complementary to recognize semantic information. 

\myPara{The complementarity between auxiliary tokens.}
To verify the complementarity between the auxiliary tokens, 
we evaluate the linear probing accuracy when combining 
two different auxiliary tokens output by the encoder. 
From the results in \figref{fig:complementary_linear_probing}, 
we can observe that 
combining two tokens can boost the performance, 
showing the different tokens are complementary. 
Meanwhile, 
we verify the complementarity between 
auxiliary tokens after the projection heads. 
Such evaluation reflects the complementarity in the pre-training pretext task. 
Specifically, taking the self-clustering based framework~\cite{caron2021emerging,zhou2021ibot} as an example, 
we combine two auxiliary tokens and feed them into the prototype layer to acquire pseudo labels. 
As shown in \figref{fig:complementary_nmi}, 
we report the normalized mutual information~(NMI) between the pseudo and true labels and
observe that different tokens are complementary. 

\begin{figure}[t]
  \centering
  \begin{overpic}[width=0.75\columnwidth]{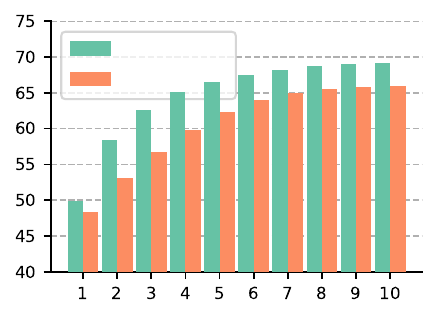}
    \put(19.0, -1){The number of auxiliary tokens}
    \put(26, 60.1){independent}
    \put(26, 53.2){shared}
    \put(-2, 31){\rotatebox{90}{NMI (\%)}}
  \end{overpic}
  \caption[complementary]{
    The complementary capabilities when combining different numbers of auxiliary tokens. 
    With ten auxiliary tokens pre-trained by our proposed \ourMthd, 
    we combine different numbers of auxiliary tokens to 
    generate pseudo labels and 
    report the normalized mutual information~(NMI) with true labels. 
    For a specific number, we evaluate every combination 
    and report the average NMI across all combinations.  
  }
  \label{fig:complementary_nmi_different_n_sharehead}
\end{figure}

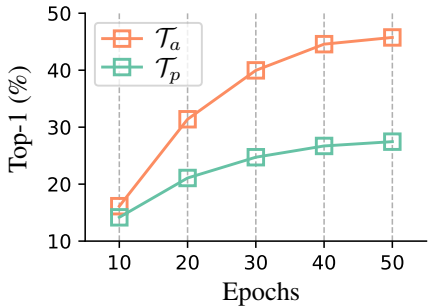
\begin{figure}[t]
  \centering
  \begin{overpic}[width=0.65\linewidth]{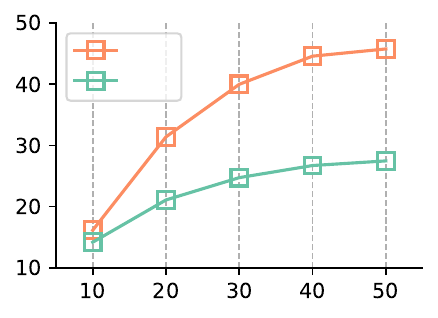}
    \put(46.5, -3){Epochs}
    \put(-6, 28){\rotatebox{90}{Top-1 (\%)}}
    \put(30, 59.2){$\mathcal{T}_a$}
    \put(30, 51.8){$\mathcal{T}_p$}
  \end{overpic}
  \caption{
    Linear probing on auxiliary CLS tokens~($\mathcal{T}_a$) and adaptively pooled tokens~($\mathcal{T}_p$)  
    when only imposing loss on the global token, 
    \ie the auxiliary tokens are not directly optimized. 
    The model is pre-trained for 50 epochs on ImageNet-1K~\cite{russakovsky2015imagenet} 
    and fine-tuned for 5 epochs. 
    The two accuracy curves are reported by averaging the accuracies of four auxiliary CLS tokens 
    and six adaptively pooled tokens, respectively. 
  }
  \label{fig:effect_of_distillation}
\end{figure}

\myPara{Impacts of the independent projection heads.}
In this work, 
we adopt independent projection heads for different auxiliary tokens. 
Because there are differences across different projection heads, 
customizing a projection head for each auxiliary token 
can better handle the unique
information of each auxiliary token. 
As shown in \figref{fig:complementary_nmi_different_n_sharehead}, 
the independent projection heads can achieve higher improvements  
when combining multiple auxiliary tokens, 
compared to using a shared projection head. 
\tabref{tab:independent_head} 
also shows that using independent heads 
leads to better performances 
on the distilled global token.

\subsection{Analysis of The Online Distillation.} 
\label{sec:analysis_distillation}

\begin{figure}[t]
  \centering
  \begin{overpic}[width=0.65\linewidth]{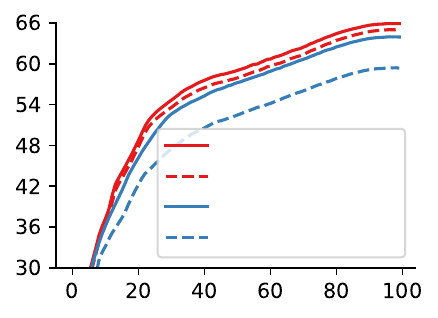}
    \put(46.0, -3){Epochs}
    \put(-6, 28.5){\rotatebox{90}{NMI~(\%)}}
    \put(50, 31){auxiliary w/ dis.}
    \put(50, 38){global w/ dis.}
    \put(50, 17){auxiliary w/o dis.}
    \put(50, 24){global w/o dis.}
  \end{overpic}
  \caption{
    Training dynamic of the normalized mutual information 
    between the true and pseudo labels with or without the online distillation~(abbreviated as dis. in the figure). 
    The pseudo labels are generated using the clustering-based self-supervised method~(DINO~\cite{caron2021emerging}+\ourMthd). 
    When without distillation, 
    we define the loss as ${\mathcal{L}}(\hat{h}_t, \tilde{h}_t)+{\mathcal{L}}(\hat{h}_c, \tilde{h}_c)$, instead of \eqref{eq:overall_loss}. 
  }
  \label{fig:necessity_of_distillation}
\end{figure}

We first verify whether the global CLS token 
can acquire the knowledge of the auxiliary tokens after the online distillation. 
\figref{fig:necessity_of_distillation} 
shows the representation quality of the global token and auxiliary tokens, respectively. 
With distillation, 
the gap between the global CLS token and auxiliary tokens 
is significantly reduced compared to without distillation, 
showing that the global CLS token learns most of the knowledge held by the auxiliary tokens.

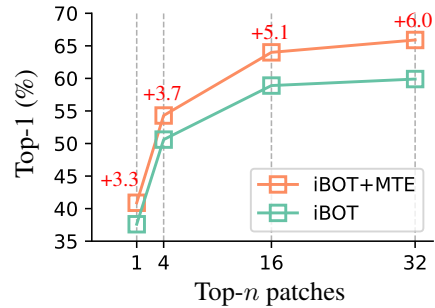
\begin{figure}[t]
  \centering
  \begin{overpic}[width=0.65\linewidth]{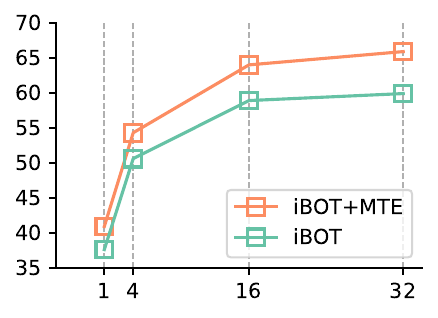}
    \put(38, -3){Top-$n$ patches}
    \put(15, 25){\footnotesize \red{+3.3}}
    \put(25.5, 46){\footnotesize \red{+3.7}}
    \put(52, 61){\footnotesize \red{+5.1}}
    \put(87, 64){\footnotesize \red{+6.0}}
    \put(-6, 27){\rotatebox{90}{Top-1 (\%)}}
  \end{overpic}
  \caption{
    The $k$-NN when using $n$ patch tokens with the top-$n$ highest self-attention scores. 
    The models are pre-trained on ImageNet-1K~\cite{russakovsky2015imagenet} for 100 epochs. 
  }
  \label{fig:part_knn}
\end{figure}

\begin{table}[t]
  \centering
  \setlength{\tabcolsep}{4.1mm}
  \caption{The effect of the independent projection heads 
  when evaluating the distilled global token.}
  \begin{tabular}{lcccc} \toprule
    Method &  $k$-NN & Linear \\ \midrule
    Baseline & 61.2 & 67.8 \\
    One shared projection head & 63.9 & 69.6 \\
    Independent projection heads & \textbf{66.6} & \textbf{70.6} \\
    \bottomrule
  \end{tabular}
  \label{tab:independent_head}
\end{table}

\myPara{Distillation benefits the auxiliary tokens.}
More importantly, we discover that 
distillation can also enhance 
auxiliary tokens, 
not only the global token. 
We assume that the enhancement of the auxiliary tokens 
is a side effect of enhancing the global token 
because the global token and the auxiliary tokens share an encoder.  
We design an experiment to verify this assumption. 
Specifically, 
we pre-train the models by only 
calculating the loss between the global tokens of teacher and student, 
\ie ${\mathcal{L}}(\hat{h}_t, \tilde{h}_t)$. 
For the auxiliary tokens, 
we freeze their randomly initialized parameters and 
remove the corresponding loss function. 
As shown in \figref{fig:effect_of_distillation}, 
even though the auxiliary tokens are not optimized explicitly, 
they are also enhanced along with the global token. 
This phenomenon shows that 
the distillation creates 
a mutually reinforcing dynamic 
between the global token and auxiliary tokens, 
resulting in a reciprocal optimizing process.

\begin{table}[t]
  \centering
  \setlength{\tabcolsep}{7.6mm}
  \caption{The $k$-NN when using CLS token or average pooled token~(GAP) for evaluation. 
  The models are pre-trained on ImageNet-1K~\cite{russakovsky2015imagenet} for 100 epochs and 
  we report the Top-1 accuracy. 
  }
  \begin{tabular}{lcccc} \toprule
    Method  & CLS & GAP \\ \midrule
    iBOT~\cite{zhou2021ibot} & 70.3 & 50.9 \\
    iBOT~\cite{zhou2021ibot}+\ourMthd & \textbf{72.4} & \textbf{57.9} \\
    \bottomrule
  \end{tabular}
    \label{tab:knn_avgpool}
\end{table}

\begin{table}[t]
  \centering
  \setlength{\tabcolsep}{2.2mm}
  \caption{Cooperation with existing model ensemble methods. 
  $\times$ 2 means that we pre-train the model twice with different random seeds 
  and then concatenate the representations of two different models for inference. 
  }
  \begin{tabular}{lcccc} \toprule
    Method & Architecture & Epochs & $k$-NN \\ \midrule
    DINO~\cite{caron2021emerging} & \multirow{3}{*}{ViT-S/16} & \multirow{3}{*}{100} & 61.2 \\
    DINO~\cite{caron2021emerging}+\ourMthd & & & 66.6 \\
    DINO~\cite{caron2021emerging}+\ourMthd~$\times$ 2 & & & \textbf{67.1} \\
    \bottomrule
  \end{tabular}
    \label{tab:ensemble}
\end{table}

\subsection{The Impacts on Patch Representations} 
\label{sec:analysis_patch}
In the proposed method, 
we acquire auxiliary tokens through pooling patch tokens. 
Thus, we investigate the impact of our method on the patch tokens. 
Specifically, 
we evaluate the $k$-NN with patch tokens 
to probe the semantic information 
instead of using the CLS token in the standard setting. 
Following~\cite{zhou2021ibot}, 
$n$ tokens with the top-$n$ highest self-attention scores are averaged for evaluation. 
As shown in \figref{fig:part_knn}, 
\ourMthd~significantly improves performance, 
especially when using more patches. 
When using all patches, 
as shown in \tabref{tab:knn_avgpool}, 
the improvement reaches 7.0\% Top-1, 
while the improvement on the CLS token is 2.1\% Top-1. 
These significant improvements show that the \ourMthd~enhances 
the high-level semantics of patches. 

\subsection{Cooperation with Model Ensemble}
In this work, 
we make a single model extract 
different auxiliary tokens 
for effective training and ensemble. 
In this section, 
we combine the proposed \ourMthd~with 
existing ensemble strategies to demonstrate its compatibility. 
Among different ensemble strategies, 
some implicit strategies, like dropout and drop-path, 
have already been integrated into a wide range of representation learning frameworks, 
including the methods in \tabref{tab:exp_knn_linear}. 
Additionally, 
we consider the explicit ensemble in \tabref{tab:ensemble} 
and further boost the performance, 
where we independently train multiple models and 
concatenate their representations for inference. 

\section{Experiments}
\label{sec:experiments} 

\subsection{Experiment Settings}

\begin{table}[t]
  \centering
  \setlength{\tabcolsep}{0.9mm}
  \caption{Experiments with different forms of loss functions 
    on the ImageNet-1K~\cite{russakovsky2015imagenet} dataset 
    without multi-crop augmentation~\cite{caron2020unsupervised}. 
    For BYOL~\cite{byol}, we implement the loss function using the codebase of DINO~\cite{caron2021emerging}.
  }
  \begin{tabular}{lcccc} \toprule
    Method & Architecture & Epochs & {$k$-NN} & {Linear} \\ \midrule
    BYOL~\cite{byol} & \tRows{ViT-S/16} & 100 & 56.1 & 65.1 \\
    BYOL~\cite{byol}+\ourMthd & & 100 & \textbf{65.5} & \textbf{70.7} \\
    \midrule
    MoCo V3~\cite{Chen_2021_ICCV} & \tRows{ViT-S/16} & 300 & - & 72.5 \\
    MoCo V3~\cite{Chen_2021_ICCV}+\ourMthd & & 300 & \textbf{67.9} & \textbf{73.3} \\
    \midrule
    DINO~\cite{caron2021emerging} & \tRows{ViT-S/16} & 100 & 61.2 & 67.8 \\
    DINO~\cite{caron2021emerging}+\ourMthd & & 100 & \textbf{66.6} & \textbf{70.6} \\
    \midrule
    iBOT~\cite{zhou2021ibot} & \tRows{ViT-S/16} & 100 & 65.2 & 71.3 \\
    iBOT~\cite{zhou2021ibot}+\ourMthd & & 100 & \textbf{68.8} & \textbf{73.1} \\
    \midrule
    HSSL~\cite{li2023enhancing} & \tRows{ViT-S/16} &100  & 67.3 & 72.6 \\
    HSSL~\cite{li2023enhancing}+\ourMthd & & 100 & \textbf{69.3} & \textbf{73.6} \\
    \midrule
    DINO~\cite{caron2021emerging} & \thRows{ViT-S/16} & 300 & 67.9 & 72.5 \\
    DINO~\cite{caron2021emerging}+\ourMthd & & 300 & \textbf{70.8} & \textbf{73.7} \\
    \textcolor{gray}{DINO~\cite{caron2021emerging}} & & \gray{800} & \gray{70.0} & \gray{73.7} \\
    \midrule
    iBOT~\cite{zhou2021ibot} & \thRows{ViT-S/16} & 300 & 70.3 & 74.8 \\
    iBOT~\cite{zhou2021ibot}+\ourMthd & & 300 & \textbf{72.4} & \textbf{75.4} \\
    \textcolor{gray}{iBOT~\cite{zhou2021ibot}} & & \gray{800} & \gray{72.4} & \gray{76.2} \\
    \midrule
    DINO~\cite{caron2021emerging} & \tRows{Swin-T} & 100 & 62.5 & 70.1 \\
    DINO~\cite{caron2021emerging}+\ourMthd & & 100 & \textbf{67.1} & \textbf{72.1} \\
    \midrule
    MoBY~\cite{xie2021moby} & \tRows{Swin-T} & 100 & - & 70.9 \\
    MoBY~\cite{xie2021moby}+\ourMthd & & 100 & \textbf{66.0} & \textbf{72.8} \\
    \midrule
    MoBY~\cite{xie2021moby} & \tRows{Swin-T} & 300 & 67.4 & 75.3 \\
    MoBY~\cite{xie2021moby}+\ourMthd & & 300 & \textbf{71.4} & \textbf{76.1} \\
    \bottomrule
  \end{tabular}
  \label{tab:exp_knn_linear}
\end{table}

\myPara{Pre-training settings.} 
When implementing our \ourMthd~with different loss functions, 
including DINO~\cite{caron2021emerging}, iBOT~\cite{zhou2021ibot}, 
MoCo~\cite{Chen_2021_CVPR}, BYOL~\cite{byol}, 
and MoBY~\cite{xie2021moby}, 
we pre-train the models on the ImageNet-1K~\cite{russakovsky2015imagenet} dataset 
and follow the settings of 
the corresponding baselines 
for a fair comparison. 
For the implementation of \ourMthd, 
the convolution used to generate adaptive pooling weights in 
\secref{sec:multi_token_define} 
comprises a $1 \times 1$ point-wise convolution and an 
$11 \times 11$ depth-wise convolution. 

\myPara{Architectures.} 
We implement our \ourMthd~upon 
ViT~\cite{dosovitskiy2020vit} and Swin~\cite{liu2021Swin}. 
For ViT, 
we pre-train the models  
using four auxiliary CLS tokens and six adaptively pooled tokens, 
with the original CLS token as the global token. 
For the Swin transformer that does not support the CLS token, 
we only use six adaptively pooled tokens, 
and the global token is an average pooled token without the adaptive weights. 
During inference, 
all auxiliary tokens are removed.

\begin{table}[t]
  \centering
  \setlength{\tabcolsep}{1.4mm}
  \caption{Experiments on the ImageNet-1K~\cite{russakovsky2015imagenet} dataset 
    with multi-crop augmentation~\cite{caron2020unsupervised}. 
    Following~\cite{zhou2021ibot}, 
    we use two large crops of $224^2$ and 
    ten small crops of $96^2$ for pre-training. 
  }
  \begin{tabular}{lcccc} \toprule
    Method & Architecture & Epochs & {$k$-NN} & {Linear} \\ \midrule
    iBOT~\cite{zhou2021ibot} & \tRows{ViT-S/16} & 100 & 71.5 & 74.4 \\
    iBOT~\cite{zhou2021ibot}+\ourMthd & & 100 & \textbf{74.1} & \textbf{76.7} \\
    \midrule
    iBOT~\cite{zhou2021ibot} & \tRows{ViT-B/16} & 100 & 74.0 & 77.8 \\
    iBOT~\cite{zhou2021ibot}+\ourMthd & & 100 & \textbf{76.0} & \textbf{78.9} \\
    \bottomrule
  \end{tabular}
    \label{tab:exp_mc_knn_linear}
\end{table}

\begin{table}[t]
  \centering
  \setlength{\tabcolsep}{0.7mm}
  \caption{Fully fine-tuning on the ImageNet-1K~\cite{russakovsky2015imagenet} dataset. 
  The ViT-S/16 and ViT-B/16 are fine-tuned for 200 and 100 epochs, respectively. 
  All models are pre-trained with the multi-crop augmentation~\cite{caron2020unsupervised}. 
  For the image size of $384^2$, 
  we further fine-tune the model for 10 epochs 
  after fine-tuning with $224^2$, 
  following~\cite{bao2021beit}. 
  }
  \begin{tabular}{lccccc} \toprule
    Method & Architecture & Epochs & Image Size & Top-1 \\ \midrule
    iBOT~\cite{zhou2021ibot} & \tRows{ViT-S/16} & 100 & $224^2$ & 81.7 \\
    iBOT~\cite{zhou2021ibot}+\ourMthd & & 100 & $224^2$ & \textbf{82.0} \\ \midrule
    iBOT~\cite{zhou2021ibot} & \tRows{ViT-B/16} & 100 & $224^2$ & 83.3 \\
    iBOT~\cite{zhou2021ibot}+\ourMthd & & 100 & $224^2$ & \textbf{83.8} \\ \midrule
    iBOT~\cite{zhou2021ibot} & \tRows{ViT-B/16} & 400 & $384^2$ & 85.0 \\
    iBOT~\cite{zhou2021ibot}+\ourMthd & & \textbf{100} & $384^2$ & \textbf{85.2} \\ 
    \bottomrule
  \end{tabular}
  \label{tab:full_finetuning}
\end{table}

\subsection{Experiment on ImageNet-1K}

After pre-training on the training set of ImageNet-1K, 
we evaluate the models on the validation set 
by $k$-NN classifier, linear probing, and fully fine-tuning, respectively. 
These results are shown in \tabref{tab:exp_knn_linear}, 
\tabref{tab:exp_mc_knn_linear}, and \tabref{tab:full_finetuning}, respectively. 

\myPara{Classification with $k$-NN and linear probing.}
Following~\cite{caron2021emerging,zhou2021ibot}, 
we report the $k$-NN Top-1 accuracies with $k$ as ten 
and the linear probing Top-1 accuracies 
by fine-tuning a linear classifier upon the frozen representations. 
In \tabref{tab:exp_knn_linear}, 
we observe that \ourMthd~consistently improves 
over different baselines. 
Compared to contrastive-based MoCo V3~\cite{Chen_2021_ICCV}, 
we achieve a 0.8\% improvement on linear probing. 
Compared to clustering-based DINO~\cite{caron2021emerging} 
and iBOT~\cite{zhou2021ibot} with 300 epochs, 
\ourMthd~improves the $k$-NN accuracy
by 2.9\% and 2.1\%, respectively, 
and improves linear probing accuracy by 1.2\% and 0.6\%, respectively. 
When using Swin transformer~\cite{liu2021Swin}, 
our \ourMthd~also improves the performance. 
For example, 
\ourMthd~advances MoBY~\cite{xie2021moby} by 
4.0\% and 0.8\% on $k$-NN and linear probing, respectively, 
verifying the extendibility of \ourMthd~to various architectures. 

Meanwhile, 
our \ourMthd~could 
achieve comparable or even better performances 
with fewer training epochs. 
Compared to DINO~\cite{caron2021emerging} with 800 epochs, 
DINO~\cite{caron2021emerging}+\ourMthd~with 300 epochs 
improves the $k$-NN accuracy by 0.8\% and 
achieves a consistent performance on linear probing. 
Compared to iBOT~\cite{zhou2021ibot} with 800 epochs, 
iBOT~\cite{caron2021emerging}+\ourMthd~with 300 epochs 
also has a consistent $k$-NN accuracy. 
These results show the capacity of the \ourMthd~to learn 
semantic representations. 

Following~\cite{caron2020unsupervised}, 
we further evaluate the effectiveness of our \ourMthd~when 
using multi-crop augmentation. 
Specifically, 
two large views of 224$\times$224 and ten small views 
of 96$\times$96 are generated. 
Due to too high computational cost, 
we pre-train the models for 100 epochs and 
the results are shown in \tabref{tab:exp_mc_knn_linear}. 
For the baseline of ViT-S/16, 
the performances of 100 epochs are from the official paper of iBOT~\cite{zhou2021ibot}. 
For ViT-B/16, we report the baseline performances by 
running the officially released codes. 
As shown in \tabref{tab:exp_mc_knn_linear}, 
our \ourMthd~improves the $k$-NN and linear probing performances. 

\begin{table}[t]
  \centering
  \setlength{\tabcolsep}{3.1mm}
  \caption{Experiments with supervised learning on ImageNet-1K~\cite{russakovsky2015imagenet}. 
  }
  \begin{tabular}{lccccc} \toprule
    Architecture & with \ourMthd & Image Size & Top-1 \\ \midrule
    \tRows{ViT-S/16~\cite{dosovitskiy2020vit}} & \xmark & $224^2$ & 79.8 \\
    & \cmark & $224^2$ & \textbf{80.4} \\\midrule
    \tRows{ViT-B/16~\cite{dosovitskiy2020vit}} & \xmark & $224^2$ & 82.3 \\
    & \cmark & $224^2$ & \textbf{82.7} \\
    \bottomrule
  \end{tabular}
  \label{tab:exp_supervised}
\end{table}

\begin{table}[t]
  \centering
  \setlength{\tabcolsep}{1.3mm}
  \caption{Transferring learning on the iNaturalist~\cite{Horn_2018_CVPR} datasets.}   
  \begin{tabular}{lccccc}
      \toprule
      Method & {Architecture} & Epoch$^\dag$ & INat$_{18}$ & INat$_{19}$ \\
      \midrule
      iBOT~\cite{zhou2021ibot} & \thRows{ViT-B/16} & 100 & 74.0 & 78.4 \\
      iBOT~\cite{zhou2021ibot}+\ourMthd & & 100 & \textbf{75.4} & \textbf{79.9} \\
      \gray{iBOT~\cite{zhou2021ibot}} & & \gray{400} & \gray{74.6} & \gray{79.6} \\
      \bottomrule
  \end{tabular}
  \label{tab:exp_transfer_cls}
\end{table}

\begin{table}[t]
  \centering
  \setlength{\tabcolsep}{1.2mm}
    \caption{Object detection and instance segmentation 
    on the COCO~\cite{lin2015microsoft} dataset.}   
    \begin{tabular}{lccccccc}
      \toprule
      Method & {Architecture} & Epoch$^\dag$ & mAP$^{b}_{}$ & mAP$^{m}_{}$ \\
      \midrule
      iBOT~\cite{zhou2021ibot} & \thRows{ViT-B/16} & 100 & 50.1 & 43.2 \\
      iBOT~\cite{zhou2021ibot}+\ourMthd & & 100 & \textbf{51.0} & \textbf{44.2} \\ 
      \gray{iBOT~\cite{zhou2021ibot}} & & \gray{400} & \gray{51.2} & \gray{44.2} \\
      \bottomrule
  \end{tabular}
  \label{tab:exp_coco_detection}
\end{table}

\begin{table}[t]
  \centering
  \setlength{\tabcolsep}{2.9mm}
  \caption{Semantic segmentation on the ADE20K~\cite{Zhou_2017_CVPR} dataset. 
  }
  \begin{tabular}{lcccc} \toprule
    Method & Architecture & Epoch$^\dag$ & mIoU \\ \midrule
    iBOT~\cite{zhou2021ibot} & \tRows{ViT-B/16} & 100 & 47.9 \\
    iBOT~\cite{zhou2021ibot}+\ourMthd & & 100 & \textbf{49.1} \\ 
    \bottomrule
  \end{tabular}
  \label{tab:exp_semantic_segmentation}
\end{table}

\myPara{Classification with fully fine-tuning.}
Apart from $k$-NN and linear probing that use frozen features, 
we also fully fine-tune the pre-trained encoders on the ImageNet-1K 
training set 
following~\cite{zhou2021ibot}. 
For fine-tuning, 
we use an AdamW~\cite{adamw} optimizer and a batch size of 1024 by default.
For ViT-S/16 and ViT-B/16, we fine-tune the models for 100 and 200 epochs, respectively. 
The learning rate is set as 2e-3 and gradually decays to 1e-6 during fine-tuning. 
And the layer-wise decay rate is 0.75 and 0.60 
for ViT-S/16 and ViT-B/16, respectively. 

As shown in \tabref{tab:full_finetuning}, 
we improve the Top-1 accuracies by 0.3\% and 0.5\%  
on ViT-S/16 and ViT-B/16, respectively. 
Following~\cite{bao2021beit}, 
we also fine-tune the models with a resolution of 384$\times$384. 
We observe that we perform better with fewer pre-training epochs, 
\ie 85.2 (100 epochs) vs 85.0 (400 epochs).

\myPara{Application on supervised learning.} 
\tabref{tab:exp_supervised} shows the supervised classification accuracy 
with and without our proposed \ourMthd. 
For example, 
we follow the training recipe in \cite{he2021masked} 
to train ViT-B/16 on ImageNet-1K~\cite{russakovsky2015imagenet} for 300 epochs 
and improve 0.4\% on Top-1 accuracy over the baseline, 
showing the extendibility of our \ourMthd~to the supervised scenario. 

\myPara{Computational costs.}
\tabref{tab:computational_cost} lists the 
computational costs during pre-training and inference. 
During pre-training, 
the auxiliary tokens, 
token enhancing module, 
and the independent projection heads 
bring some extra computational budget. 
When inference, 
they are discarded without additional inference costs. 

\begin{table}[t]
  \centering
  \setlength{\tabcolsep}{1.4mm}
  \caption{Computational costs during pre-training and inference. 
  We report the floating-point operations per second~(FLOPS) 
  with an input image of $224 \times 224$.}  
  \begin{tabular}{lccccccc}
      \toprule
      & {Architecture} & {Pre-training} & {Inference} \\
      \midrule
      iBOT~\cite{zhou2021ibot} & \tRows{ViT-S/16} & 6.1G & 4.6G \\
      iBOT~\cite{zhou2021ibot}+\ourMthd & & 6.4G & 4.6G \\
      \bottomrule
  \end{tabular} 
  \label{tab:computational_cost}
\end{table}

\subsection{Transferring Learning}

\myPara{Image classification.}
Besides the ImageNet~\cite{russakovsky2015imagenet}, 
we also transfer the pre-trained models to other datasets
whose domains differ from the ImageNet. 
Specifically, we fine-tune the pre-trained models for 360 epochs on 
iNaturalist 2018~\cite{Horn_2018_CVPR} and iNaturalist 2019~\cite{Horn_2018_CVPR} datasets, 
which contain 8,142 and 1,010 categories, respectively. 
As shown in \tabref{tab:exp_transfer_cls}, 
\ourMthd~consistently performs better 
and even outperforms the baseline of 400 epochs with only 100 epochs, 
demonstrating stronger transferability.

\myPara{Object detection and instance segmentation.}
We also evaluate the pre-trained models on downstream tasks with 
dense predictions. 
For instance segmentation, 
we use Cascaded Mask RCNN~\cite{Cai_2018_CVPR} as the detector and 
fine-tune the models on the COCO train2017 split~\cite{lin2015microsoft} 
for 12 epochs, following~\cite{zhou2021ibot}.

For evaluation, 
we report the mean average precision of 
bounding box (mAP$^{b}_{}$) and instance
segmentation (mAP$^{m}_{}$) on COCO val2017 split, 
as shown in \tabref{tab:exp_coco_detection}. 
Compared to iBOT~\cite{zhou2021ibot}, 
we improve the mAP$^{b}_{}$ and mAP$^{m}_{}$ 
by 0.9\% and 1.0\%, respectively. 
Meanwhile, compared to iBOT~\cite{zhou2021ibot} of 400 epochs, 
we can achieve considerable mAP$^{m}_{}$ 
with only 100 epochs.

\myPara{Semantic segmentation.}
We conduct the experiments on the ADE20K~\cite{Zhou_2017_CVPR} dataset 
and adopt UperNet~\cite{Xiao_2018_ECCV} as the segmentation model. 
An auxiliary segmentation head
is applied after the eighth block of ViT. 
Following~\cite{zhou2021ibot}, 
we fine-tune the models with 160,000 iterations and a total 
batch size of 16. 
The learning rate is initially set as 8e-5 and 
has a layer-wise decay rate of 0.9. 
For data augmentations, 
the images are randomly scaled with a ratio between
0.5 and 2.0 and randomly cropped to 512 $\times$ 512.

\tabref{tab:exp_semantic_segmentation} reports the 
mean intersection over union (mIoU) on the validation set of ADE20K. 
\ourMthd~improves the mIoU 
by 1.2\% over the iBOT~\cite{zhou2021ibot}. 
These results verify that \ourMthd~can benefit the vision tasks 
with dense predictions.

\begin{table}[t]
  \centering
  \setlength{\tabcolsep}{5.2mm}
  \caption{Ablation studies of the proposed \ourMthd. 
  $\hat{\mathcal{T}_a}$ and $\hat{\mathcal{T}_p}$ mean the 
  auxiliary CLS tokens and adaptively pooled tokens, respectively. 
  TEN means the token enhancing module.}
  \begin{tabular}{lcccc}	
    \toprule
    $\hat{\mathcal{T}_a}$ & $\hat{\mathcal{T}_p}$ & TEN & Top-1 & Top-5 \\ 
    \midrule
    \xmark & \xmark & \xmark & 67.5 & 84.4 \\
    \cmark & \xmark & \xmark & 69.7 & 85.9 \\
    \cmark & \xmark & \cmark & 70.7 & 86.5 \\
    \xmark & \cmark & \xmark & 73.0 & 87.4 \\
    \cmark & \cmark & \cmark & \textbf{73.8} & \textbf{87.8} \\
    \bottomrule
  \end{tabular}
  \label{tab:ablation}
\end{table}

\begin{figure}[t]
  \centering
  \begin{subfigure}[t]{0.48\columnwidth}
      \begin{overpic}[width=1.0\columnwidth]{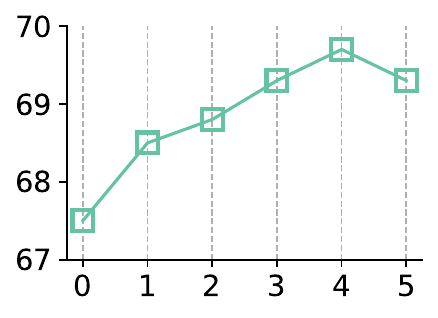}
      \put(22.0, -4){\small Auxiliary CLS tokens}
      \put(-4.0, 30.5){\rotatebox{90}{\small Top-1}}
      \end{overpic}
  \end{subfigure}
  \begin{subfigure}[t]{0.48\columnwidth}
      \begin{overpic}[width=1.0\columnwidth]{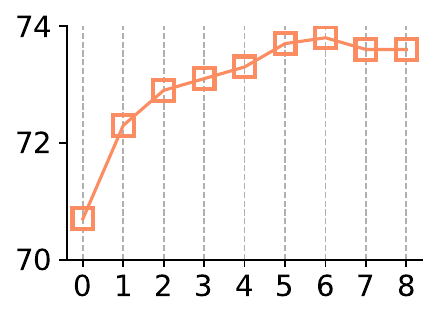}
      \put(15.2, -4){\small Adaptively pooled tokens}
      \put(-4.0, 30.5){\rotatebox{90}{\small Top-1}}
      \end{overpic}
  \end{subfigure}
  \caption{
  The $k$-NN performance when using different numbers of auxiliary tokens. 
  For all experiments of adaptively pooled tokens, 
  four auxiliary CLS tokens are also used. 
  }
  \label{fig:different_number_of_auxiliary_tokens}
\end{figure}

\begin{table}[t]
  \centering
  \setlength{\tabcolsep}{5.8mm}
  \caption{Effect of the independent classifiers for 
    modeling different auxiliary tokens in supervised learning. 
    The models are trained for 100 epochs, 
    and the 
    other settings are the same as \tabref{tab:exp_supervised}.
  }
  \begin{tabular}{lccc}	
    \toprule
    & Top-1 & Top-5 \\ 
  \midrule
    shared classifiers & 81.2 & 95.6 \\
    independent classifiers & \textbf{81.6} & \textbf{95.8} \\
  \bottomrule
  \end{tabular}
  \label{tab:sup_ablation_independent_classifiers}
\end{table}

\begin{table}[t]
  \centering
  \setlength{\tabcolsep}{7.2mm}
  \caption{Effects of different pooling operations. }
  \begin{tabular}{lccc}
      \toprule
      pooling function & Top-1 & Top-5 \\ 
      \midrule
      average & 71.0 & 86.5 \\
      max & 71.2 & 86.5 \\
      adaptive & \textbf{72.3} & \textbf{87.3} \\
      \bottomrule
  \end{tabular}
  \label{tab:pooling_operation}
\end{table}

\subsection{Ablation Studies}
\label{sec:ablation}

In this section, 
we analyze the effect of the proposed Multi-Token Enhancing 
using DINO~\cite{caron2021emerging} as the baseline. 
To save computational costs, 
we use a dataset 
containing 300 categories~\cite{gao2022luss} of ImageNet-1K~\cite{russakovsky2015imagenet} 
and pre-train the models for 100 epochs with a batch size of 256.

\myPara{Ablation study on the auxiliary tokens.}
In \secref{sec:multi_token_define}, 
we design two types of auxiliary tokens to enhance 
the instance discrimination. 
As shown in \tabref{tab:ablation}, 
both the auxiliary CLS tokens and adaptively pooled tokens 
significantly improve the performance. 
Meanwhile, 
the token enhancing module, 
which enhances the auxiliary CLS tokens and thus provides the global token with 
a better distillation teacher, 
also improves 1.0\% Top-1 accuracy. 
These results demonstrate the effectiveness of our proposed method. 

\figref{fig:different_number_of_auxiliary_tokens} further shows the impacts 
of the number of auxiliary tokens. 
We can observe that more auxiliary tokens 
can bring more significant improvements, 
and the improvement reaches saturation 
with four auxiliary CLS tokens and six adaptively pooled tokens. 
Thus, by default, we set the number of auxiliary CLS tokens and adaptively pooled tokens 
as four and six. 

\myPara{Independent heads in supervised learning.} 
In \secref{sec:multi_token_ssl}, 
we use independent projection heads to process 
each auxiliary token 
and verify the advantages in \secref{sec:analysis_difference}. 
We also verify the effect of independent classifiers in supervised learning, 
as shown in \tabref{tab:sup_ablation_independent_classifiers}, 
where using independent classifiers leads to a better performance than a shared classifier. 

\myPara{Effect of the pooling operation.}
In \secref{sec:multi_token_define}, 
we adopt adaptive weights to obtain pooled tokens. 
\tabref{tab:pooling_operation} compares the effects of different types of pooling, 
where adaptive pooling outperforms 
both average and max pooling. 
On the one hand, 
average pooling mixes different foreground and background objects, 
resulting in much noise and missing discriminative object information, and 
max pooling loses much helpful information. 
In contrast, 
adaptive pooling can focus on important regions. 
Meanwhile, 
adaptive pooling enables us to generate multiple tokens through different pooling weights. 
However, the average and max pooling can only produce one token, limiting their potential. 

\myPara{Attention masking.}
In \secref{sec:online_distillation}, 
we mask the attention between the global token and the auxiliary tokens, 
to prevent the global CLS token from relying on the auxiliary tokens to understand image semantics. 
As shown in \tabref{tab:ablation_cut_off_attention}, 
this operation avoids sacrificing performance after removing the auxiliary tokens.

\begin{table}[t]
  \centering
  \setlength{\tabcolsep}{1.3mm}
  \caption{Effect of masking the attention between the global token and the auxiliary tokens during pre-training. 
  During inference, the \cmark~and \xmark~means that the auxiliary tokens remain and are removed, respectively.}
  \begin{tabular}{lccc}
      \toprule
      & auxiliary tokens & Top-1 & Top-5 \\ 
      \midrule
      not cutting off attention & \cmark & 73.6 & 87.4 \\
      not cutting off attention & \xmark & 71.0 & 86.2 \\
      \midrule
      cutting off attention & \cmark & \textbf{73.8} & \textbf{87.8} \\
      cutting off attention & \xmark & \textbf{73.8} & \textbf{87.8} \\
      \bottomrule
  \end{tabular}
  \label{tab:ablation_cut_off_attention}
\end{table}

\section{Conclusion}
This paper 
proposes Multi-Token Enhancing~(\ourMthd) to 
enhance vision representation learning, 
including self-supervised and supervised learning. 
Specifically, 
we extract multiple auxiliary tokens from a single model, 
where the auxiliary tokens can extract different information and 
thus complement each other to enhance the representations. 
Moreover, we distill the auxiliary tokens into another global token in an online way. 
After pre-training, 
the global token has acquired a strong capacity to understand images and 
is used for inference, 
and the auxiliary tokens are removed to avoid sacrificing inference speed.
We show that the proposed \ourMthd~is compatible with various loss functions of 
self-supervised learning and can be implemented in 
different architectures. 
Across both image-level and dense-level downstream tasks, 
\eg image classification, semantic segmentation, and instance segmentation, 
\ourMthd~consistently improves performance.

{
\small
\bibliographystyle{ieeenat_fullname}
\bibliography{egbib}

\begin{thebibliography}{68}
\providecommand{\natexlab}[1]{#1}
\providecommand{\url}[1]{\texttt{#1}}
\expandafter\ifx\csname urlstyle\endcsname\relax
  \providecommand{\doi}[1]{doi: #1}\else
  \providecommand{\doi}{doi: \begingroup \urlstyle{rm}\Url}\fi

\bibitem[Asano et~al.(2020)Asano, Rupprecht, and Vedaldi]{asano2020self}
Yuki~M. Asano, Christian Rupprecht, and Andrea Vedaldi.
\newblock Self-labelling via simultaneous clustering and representation
  learning.
\newblock In \emph{ICLR}, 2020.

\bibitem[Bao et~al.(2022)Bao, Dong, Piao, and Wei]{bao2021beit}
Hangbo Bao, Li Dong, Songhao Piao, and Furu Wei.
\newblock {BE}it: {BERT} pre-training of image transformers.
\newblock In \emph{ICLR}, 2022.

\bibitem[Cai and Vasconcelos(2018)]{Cai_2018_CVPR}
Zhaowei Cai and Nuno Vasconcelos.
\newblock Cascade r-cnn: Delving into high quality object detection.
\newblock In \emph{CVPR}, 2018.

\bibitem[Caron et~al.(2020)Caron, Misra, Mairal, Goyal, Bojanowski, and
  Joulin]{caron2020unsupervised}
Mathilde Caron, Ishan Misra, Julien Mairal, Priya Goyal, Piotr Bojanowski, and
  Armand Joulin.
\newblock Unsupervised learning of visual features by contrasting cluster
  assignments.
\newblock In \emph{NeurIPS}, 2020.

\bibitem[Caron et~al.(2021)Caron, Touvron, Misra, J\'egou, Mairal, Bojanowski,
  and Joulin]{caron2021emerging}
Mathilde Caron, Hugo Touvron, Ishan Misra, Herv\'e J\'egou, Julien Mairal,
  Piotr Bojanowski, and Armand Joulin.
\newblock Emerging properties in self-supervised vision transformers.
\newblock In \emph{ICCV}, 2021.

\bibitem[Chen et~al.(2023)Chen, Liu, Hong, Xu, Li, and Yeung]{Chen_2023_CVPR}
Kai Chen, Zhili Liu, Lanqing Hong, Hang Xu, Zhenguo Li, and Dit-Yan Yeung.
\newblock Mixed autoencoder for self-supervised visual representation learning.
\newblock In \emph{CVPR}, pages 22742--22751, 2023.

\bibitem[Chen et~al.(2020)Chen, Kornblith, Norouzi, and Hinton]{chen2020simple}
Ting Chen, Simon Kornblith, Mohammad Norouzi, and Geoffrey Hinton.
\newblock A simple framework for contrastive learning of visual
  representations.
\newblock In \emph{ICML}, 2020.

\bibitem[Chen and He(2021)]{Chen_2021_CVPR}
Xinlei Chen and Kaiming He.
\newblock Exploring simple siamese representation learning.
\newblock In \emph{CVPR}, 2021.

\bibitem[Chen et~al.(2021)Chen, Xie, and He]{Chen_2021_ICCV}
Xinlei Chen, Saining Xie, and Kaiming He.
\newblock An empirical study of training self-supervised vision transformers.
\newblock In \emph{ICCV}, 2021.

\bibitem[D'Ascoli et~al.(2021)D'Ascoli, Touvron, Leavitt, Morcos, Biroli, and
  Sagun]{convit}
Stéphane D'Ascoli, Hugo Touvron, Matthew~L Leavitt, Ari~S Morcos, Giulio
  Biroli, and Levent Sagun.
\newblock Convit: Improving vision transformers with soft convolutional
  inductive biases.
\newblock In \emph{ICML}, 2021.

\bibitem[Dietterich(2000)]{dietterich2000ensemble}
Thomas~G Dietterich.
\newblock Ensemble methods in machine learning.
\newblock In \emph{International workshop on multiple classifier systems},
  pages 1--15. Springer, 2000.

\bibitem[Dosovitskiy et~al.(2021)Dosovitskiy, Beyer, Kolesnikov, Weissenborn,
  Zhai, Unterthiner, Dehghani, Minderer, Heigold, Gelly, Uszkoreit, and
  Houlsby]{dosovitskiy2020vit}
Alexey Dosovitskiy, Lucas Beyer, Alexander Kolesnikov, Dirk Weissenborn,
  Xiaohua Zhai, Thomas Unterthiner, Mostafa Dehghani, Matthias Minderer, Georg
  Heigold, Sylvain Gelly, Jakob Uszkoreit, and Neil Houlsby.
\newblock An image is worth 16x16 words: Transformers for image recognition at
  scale.
\newblock In \emph{ICLR}, 2021.

\bibitem[Ermolov et~al.(2021)Ermolov, Siarohin, Sangineto, and
  Sebe]{ermolov2021whitening}
Aleksandr Ermolov, Aliaksandr Siarohin, Enver Sangineto, and Nicu Sebe.
\newblock Whitening for self-supervised representation learning.
\newblock In \emph{ICML}, 2021.

\bibitem[Fang et~al.(2021)Fang, Wang, Wang, Zhang, Yang, and Liu]{fang2021seed}
Zhiyuan Fang, Jianfeng Wang, Lijuan Wang, Lei Zhang, Yezhou Yang, and Zicheng
  Liu.
\newblock Seed: Self-supervised distillation for visual representation.
\newblock In \emph{ICLR}, 2021.

\bibitem[Feng and Zhang(2023)]{Feng_2023_CVPR}
Zhanzhou Feng and Shiliang Zhang.
\newblock Evolved part masking for self-supervised learning.
\newblock In \emph{CVPR}, pages 10386--10395, 2023.

\bibitem[Ganaie et~al.(2022)Ganaie, Hu, Malik, Tanveer, and
  Suganthan]{GANAIE2022105151}
M.A. Ganaie, Minghui Hu, A.K. Malik, M. Tanveer, and P.N. Suganthan.
\newblock Ensemble deep learning: A review.
\newblock \emph{Engineering Applications of Artificial Intelligence},
  115:\penalty0 105151, 2022.

\bibitem[Gao et~al.(2022)Gao, Li, Yang, Cheng, Han, and Torr]{gao2022luss}
Shanghua Gao, Zhong-Yu Li, Ming-Hsuan Yang, Ming-Ming Cheng, Junwei Han, and
  Philip Torr.
\newblock Large-scale unsupervised semantic segmentation.
\newblock \emph{IEEE TPAMI}, 2022.

\bibitem[Grill et~al.(2020)Grill, Strub, Altché, Tallec, Richemond,
  Buchatskaya, Doersch, Ávila Pires, Guo, Azar, Piot, Kavukcuoglu, Munos, and
  Valko]{byol}
Jean-Bastien Grill, Florian Strub, Florent Altché, Corentin Tallec, Pierre~H.
  Richemond, Elena Buchatskaya, Carl Doersch, Bernardo Ávila Pires, Zhaohan
  Guo, Mohammad~Gheshlaghi Azar, Bilal Piot, Koray Kavukcuoglu, Rémi Munos,
  and Michal Valko.
\newblock Bootstrap your own latent - a new approach to self-supervised
  learning.
\newblock In \emph{NeurIPS}, 2020.

\bibitem[Guo et~al.(2022)Guo, Lu, Liu, Cheng, and Hu]{guo2022visual}
Meng-Hao Guo, Cheng-Ze Lu, Zheng-Ning Liu, Ming-Ming Cheng, and Shi-Min Hu.
\newblock Visual attention network.
\newblock \emph{arXiv preprint arXiv:2202.09741}, 2022.

\bibitem[Hansen and Salamon(1990)]{58871}
L.K. Hansen and P. Salamon.
\newblock Neural network ensembles.
\newblock \emph{IEEE TPAMI}, 12\penalty0 (10):\penalty0 993--1001, 1990.

\bibitem[Havasi et~al.(2021)Havasi, Jenatton, Fort, Liu, Snoek,
  Lakshminarayanan, Dai, and Tran]{havasi2021training}
Marton Havasi, Rodolphe Jenatton, Stanislav Fort, Jeremiah~Zhe Liu, Jasper
  Snoek, Balaji Lakshminarayanan, Andrew~Mingbo Dai, and Dustin Tran.
\newblock Training independent subnetworks for robust prediction.
\newblock In \emph{ICLR}, 2021.

\bibitem[He et~al.(2016)He, Zhang, Ren, and Sun]{he2016deep}
Kaiming He, Xiangyu Zhang, Shaoqing Ren, and Jian Sun.
\newblock Deep residual learning for image recognition.
\newblock In \emph{CVPR}, 2016.

\bibitem[He et~al.(2020)He, Fan, Wu, Xie, and Girshick]{He_2020_CVPR}
Kaiming He, Haoqi Fan, Yuxin Wu, Saining Xie, and Ross Girshick.
\newblock Momentum contrast for unsupervised visual representation learning.
\newblock In \emph{CVPR}, 2020.

\bibitem[He et~al.(2022)He, Chen, Xie, Li, Doll\'ar, and
  Girshick]{he2021masked}
Kaiming He, Xinlei Chen, Saining Xie, Yanghao Li, Piotr Doll\'ar, and Ross
  Girshick.
\newblock Masked autoencoders are scalable vision learners.
\newblock In \emph{CVPR}, 2022.

\bibitem[Hinton et~al.(2015)Hinton, Vinyals, and Dean]{hinton2015distilling}
Geoffrey Hinton, Oriol Vinyals, and Jeff Dean.
\newblock Distilling the knowledge in a neural network.
\newblock \emph{arXiv preprint arXiv:1503.02531}, 2015.

\bibitem[Hou et~al.(2022)Hou, Lu, Cheng, and Feng]{hou2022conv2former}
Qibin Hou, Cheng-Ze Lu, Ming-Ming Cheng, and Jiashi Feng.
\newblock Conv2former: A simple transformer-style convnet for visual
  recognition.
\newblock \emph{arXiv preprint arXiv:2211.11943}, 2022.

\bibitem[Izmailov et~al.(2018)Izmailov, Podoprikhin, Garipov, Vetrov, and
  Wilson]{izmailov2018averaging}
Pavel Izmailov, Dmitrii Podoprikhin, Timur Garipov, Dmitry Vetrov, and
  Andrew~Gordon Wilson.
\newblock Averaging weights leads to wider optima and better generalization.
\newblock \emph{arXiv preprint arXiv:1803.05407}, 2018.

\bibitem[Kornblith et~al.(2019)Kornblith, Norouzi, Lee, and
  Hinton]{kornblith2019similarity}
Simon Kornblith, Mohammad Norouzi, Honglak Lee, and Geoffrey Hinton.
\newblock Similarity of neural network representations revisited.
\newblock In \emph{ICML}, 2019.

\bibitem[Larsson et~al.(2017)Larsson, Maire, and
  Shakhnarovich]{larsson2017fractalnet}
Gustav Larsson, Michael Maire, and Gregory Shakhnarovich.
\newblock Fractalnet: Ultra-deep neural networks without residuals.
\newblock In \emph{ICLR}, 2017.

\bibitem[Lee et~al.(2015)Lee, Purushwalkam, Cogswell, Crandall, and
  Batra]{lee2015m}
Stefan Lee, Senthil Purushwalkam, Michael Cogswell, David Crandall, and Dhruv
  Batra.
\newblock Why m heads are better than one: Training a diverse ensemble of deep
  networks.
\newblock \emph{arXiv preprint arXiv:1511.06314}, 2015.

\bibitem[Li et~al.(2023{\natexlab{a}})Li, Gao, and Cheng]{li2023sere}
Zhong-Yu Li, Shanghua Gao, and Ming-Ming Cheng.
\newblock Sere: Exploring feature self-relation for self-supervised
  transformer.
\newblock \emph{IEEE TPAMI}, 2023{\natexlab{a}}.

\bibitem[Li et~al.(2023{\natexlab{b}})Li, Yin, Gao, Liu, Liu, and
  Cheng]{li2023enhancing}
Zhong-Yu Li, Bo-Wen Yin, Shanghua Gao, Yongxiang Liu, Li Liu, and Ming-Ming
  Cheng.
\newblock Enhancing representations through heterogeneous self-supervised
  learning.
\newblock \emph{arXiv preprint arXiv:2310.05108}, 2023{\natexlab{b}}.

\bibitem[Lin et~al.(2014)Lin, Maire, Belongie, Hays, Perona, Ramanan,
  Doll{\'a}r, and Zitnick]{lin2015microsoft}
Tsung-Yi Lin, Michael Maire, Serge Belongie, James Hays, Pietro Perona, Deva
  Ramanan, Piotr Doll{\'a}r, and C~Lawrence Zitnick.
\newblock Microsoft coco: Common objects in context.
\newblock In \emph{ECCV}, 2014.

\bibitem[Liu et~al.(2021)Liu, Lin, Cao, Hu, Wei, Zhang, Lin, and
  Guo]{liu2021Swin}
Ze Liu, Yutong Lin, Yue Cao, Han Hu, Yixuan Wei, Zheng Zhang, Stephen Lin, and
  Baining Guo.
\newblock Swin transformer: Hierarchical vision transformer using shifted
  windows.
\newblock \emph{ICCV}, 2021.

\bibitem[Liu et~al.(2022)Liu, Mao, Wu, Feichtenhofer, Darrell, and
  Xie]{liu2022convnet}
Zhuang Liu, Hanzi Mao, Chao-Yuan Wu, Christoph Feichtenhofer, Trevor Darrell,
  and Saining Xie.
\newblock A convnet for the 2020s.
\newblock \emph{CVPR}, 2022.

\bibitem[Loshchilov and Hutter(2019)]{adamw}
Ilya Loshchilov and Frank Hutter.
\newblock Decoupled weight decay regularization.
\newblock In \emph{ICLR}, 2019.

\bibitem[Nakamura et~al.(2023)Nakamura, Okada, and
  Taniguchi]{Nakamura_2023_ICCV}
Hiroki Nakamura, Masashi Okada, and Tadahiro Taniguchi.
\newblock Representation uncertainty in self-supervised learning as variational
  inference.
\newblock In \emph{ICCV}, pages 16484--16493, 2023.

\bibitem[Navaneet et~al.(2021)Navaneet, Koohpayegani, Tejankar, and
  Pirsiavash]{navaneet2021simreg}
K~L Navaneet, Soroush~Abbasi Koohpayegani, Ajinkya Tejankar, and Hamed
  Pirsiavash.
\newblock Simreg: Regression as a simple yet effective tool for self-supervised
  knowledge distillation.
\newblock In \emph{BMVC}, 2021.

\bibitem[Oord et~al.(2018)Oord, Li, and Vinyals]{oord2018representation}
Aaron van~den Oord, Yazhe Li, and Oriol Vinyals.
\newblock Representation learning with contrastive predictive coding.
\newblock \emph{arXiv preprint arXiv:1807.03748}, 2018.

\bibitem[Russakovsky et~al.(2015)Russakovsky, Deng, Su, Krause, Satheesh, Ma,
  Huang, Karpathy, Khosla, Bernstein, et~al.]{russakovsky2015imagenet}
Olga Russakovsky, Jia Deng, Hao Su, Jonathan Krause, Sanjeev Satheesh, Sean Ma,
  Zhiheng Huang, Andrej Karpathy, Aditya Khosla, Michael Bernstein, et~al.
\newblock Imagenet large scale visual recognition challenge.
\newblock \emph{IJCV}, 115\penalty0 (3):\penalty0 211--252, 2015.

\bibitem[Shu et~al.(2023)Shu, van~den Hengel, and Liu]{Shu_2023_CVPR}
Yangyang Shu, Anton van~den Hengel, and Lingqiao Liu.
\newblock Learning common rationale to improve self-supervised representation
  for fine-grained visual recognition problems.
\newblock In \emph{CVPR}, pages 11392--11401, 2023.

\bibitem[Simpson et~al.(2022)Simpson, Vicente, and Campbell]{Simpson_2022_CVPR}
Ivor J.~A. Simpson, Sara Vicente, and Neill D.~F. Campbell.
\newblock Learning structured gaussians to approximate deep ensembles.
\newblock In \emph{Proceedings of the IEEE/CVF Conference on Computer Vision
  and Pattern Recognition (CVPR)}, pages 366--374, 2022.

\bibitem[Song et~al.(2023{\natexlab{a}})Song, Xie, Zhang, and
  Luo]{song2023multimode}
Kaiyou Song, Jin Xie, Shan Zhang, and Zimeng Luo.
\newblock Multi-mode online knowledge distillation for self-supervised visual
  representation learning.
\newblock In \emph{CVPR}, 2023{\natexlab{a}}.

\bibitem[Song et~al.(2023{\natexlab{b}})Song, Zhang, Luo, Wang, and
  Xie]{Song_2023_ICCV}
Kaiyou Song, Shan Zhang, Zimeng Luo, Tong Wang, and Jin Xie.
\newblock Semantics-consistent feature search for self-supervised visual
  representation learning.
\newblock In \emph{ICCV}, pages 16099--16108, 2023{\natexlab{b}}.

\bibitem[Srivastava et~al.(2014)Srivastava, Hinton, Krizhevsky, Sutskever, and
  Salakhutdinov]{Dropout}
Nitish Srivastava, Geoffrey Hinton, Alex Krizhevsky, Ilya Sutskever, and Ruslan
  Salakhutdinov.
\newblock Dropout: A simple way to prevent neural networks from overfitting.
\newblock \emph{Journal of Machine Learning Research}, 15\penalty0
  (56):\penalty0 1929--1958, 2014.

\bibitem[Tao et~al.(2023)Tao, Zhu, Su, Huang, Li, Zhou, Qiao, Wang, and
  Dai]{Tao_2023_CVPR}
Chenxin Tao, Xizhou Zhu, Weijie Su, Gao Huang, Bin Li, Jie Zhou, Yu Qiao,
  Xiaogang Wang, and Jifeng Dai.
\newblock Siamese image modeling for self-supervised vision representation
  learning.
\newblock In \emph{CVPR}, pages 2132--2141, 2023.

\bibitem[Tian et~al.(2020)Tian, Chen, and Ganguli]{DynamicsContrastive}
Yuandong Tian, Xinlei Chen, and Surya Ganguli.
\newblock Understanding self-supervised learning dynamics without contrastive
  pairs.
\newblock In \emph{ICML}, 2020.

\bibitem[Van~Horn et~al.(2018)Van~Horn, Mac~Aodha, Song, Cui, Sun, Shepard,
  Adam, Perona, and Belongie]{Horn_2018_CVPR}
Grant Van~Horn, Oisin Mac~Aodha, Yang Song, Yin Cui, Chen Sun, Alex Shepard,
  Hartwig Adam, Pietro Perona, and Serge Belongie.
\newblock The inaturalist species classification and detection dataset.
\newblock In \emph{CVPR}, 2018.

\bibitem[Wang et~al.(2021{\natexlab{a}})Wang, Xie, Li, Fan, Song, Liang, Lu,
  Luo, and Shao]{wang2021pyramid}
Wenhai Wang, Enze Xie, Xiang Li, Deng-Ping Fan, Kaitao Song, Ding Liang, Tong
  Lu, Ping Luo, and Ling Shao.
\newblock Pyramid vision transformer: A versatile backbone for dense prediction
  without convolutions.
\newblock In \emph{ICCV}, pages 568--578, 2021{\natexlab{a}}.

\bibitem[Wang et~al.(2021{\natexlab{b}})Wang, Zhang, Shen, Kong, and
  Li]{wang2020DenseCL}
Xinlong Wang, Rufeng Zhang, Chunhua Shen, Tao Kong, and Lei Li.
\newblock Dense contrastive learning for self-supervised visual pre-training.
\newblock In \emph{CVPR}, 2021{\natexlab{b}}.

\bibitem[Wei et~al.(2021)Wei, Fan, Xie, Wu, Yuille, and
  Feichtenhofer]{cwei2021}
Chen Wei, Haoqi Fan, Saining Xie, Chao-Yuan Wu, Alan Yuille, and Christoph
  Feichtenhofer.
\newblock Masked feature prediction for self-supervised visual pre-training.
\newblock \emph{arXiv preprint arXiv:2112.09133}, 2021.

\bibitem[Wen et~al.(2020)Wen, Tran, and Ba]{Wen2020BatchEnsemble}
Yeming Wen, Dustin Tran, and Jimmy Ba.
\newblock Batchensemble: an alternative approach to efficient ensemble and
  lifelong learning.
\newblock In \emph{ICLR}, 2020.

\bibitem[Woo et~al.(2023)Woo, Debnath, Hu, Chen, Liu, Kweon, and
  Xie]{Woo2023ConvNeXtV2}
Sanghyun Woo, Shoubhik Debnath, Ronghang Hu, Xinlei Chen, Zhuang Liu, In~So
  Kweon, and Saining Xie.
\newblock Convnext v2: Co-designing and scaling convnets with masked
  autoencoders.
\newblock In \emph{CVPR}, pages 16133--16142, 2023.

\bibitem[Wu et~al.(2021)Wu, Xiao, Codella, Liu, Dai, Yuan, and
  Zhang]{Wu_2021_ICCV}
Haiping Wu, Bin Xiao, Noel Codella, Mengchen Liu, Xiyang Dai, Lu Yuan, and Lei
  Zhang.
\newblock Cvt: Introducing convolutions to vision transformers.
\newblock In \emph{ICCV}, 2021.

\bibitem[Wu et~al.(2022)Wu, Liu, Zhan, and Cheng]{wu2022p2t}
Yu-Huan Wu, Yun Liu, Xin Zhan, and Ming-Ming Cheng.
\newblock {P2T}: Pyramid pooling transformer for scene understanding.
\newblock \emph{IEEE TPAMI}, 2022.

\bibitem[Xiao et~al.(2018)Xiao, Liu, Zhou, Jiang, and Sun]{Xiao_2018_ECCV}
Tete Xiao, Yingcheng Liu, Bolei Zhou, Yuning Jiang, and Jian Sun.
\newblock Unified perceptual parsing for scene understanding.
\newblock In \emph{ECCV}, 2018.

\bibitem[Xie et~al.(2021{\natexlab{a}})Xie, Ding, Wang, Zhan, Xu, Sun, Li, and
  Luo]{Xie_2021_ICCV}
Enze Xie, Jian Ding, Wenhai Wang, Xiaohang Zhan, Hang Xu, Peize Sun, Zhenguo
  Li, and Ping Luo.
\newblock Detco: Unsupervised contrastive learning for object detection.
\newblock In \emph{ICCV}, 2021{\natexlab{a}}.

\bibitem[Xie et~al.(2021{\natexlab{b}})Xie, Lin, Yao, Zhang, Dai, Cao, and
  Hu]{xie2021moby}
Zhenda Xie, Yutong Lin, Zhuliang Yao, Zheng Zhang, Qi Dai, Yue Cao, and Han Hu.
\newblock Self-supervised learning with swin transformers.
\newblock \emph{arXiv preprint arXiv:2105.04553}, 2021{\natexlab{b}}.

\bibitem[Xie et~al.(2021{\natexlab{c}})Xie, Lin, Zhang, Cao, Lin, and
  Hu]{Xie_2021_CVPR}
Zhenda Xie, Yutong Lin, Zheng Zhang, Yue Cao, Stephen Lin, and Han Hu.
\newblock Propagate yourself: Exploring pixel-level consistency for
  unsupervised visual representation learning.
\newblock In \emph{CVPR}, 2021{\natexlab{c}}.

\bibitem[Xie et~al.(2022)Xie, Zhang, Cao, Lin, Bao, Yao, Dai, and
  Hu]{Xie_2022_CVPR}
Zhenda Xie, Zheng Zhang, Yue Cao, Yutong Lin, Jianmin Bao, Zhuliang Yao, Qi
  Dai, and Han Hu.
\newblock Simmim: A simple framework for masked image modeling.
\newblock In \emph{CVPR}, pages 9653--9663, 2022.

\bibitem[Xu et~al.(2022)Xu, Fang, ZHANG, Xie, Wang, Dai, Xiong, and
  Tian]{xu2022bag}
Haohang Xu, Jiemin Fang, XIAOPENG ZHANG, Lingxi Xie, Xinggang Wang, Wenrui Dai,
  Hongkai Xiong, and Qi Tian.
\newblock Bag of instances aggregation boosts self-supervised distillation.
\newblock In \emph{ICLR}, 2022.

\bibitem[Yang et~al.(2016)Yang, Parikh, and Batra]{yangCVPR2016joint}
Jianwei Yang, Devi Parikh, and Dhruv Batra.
\newblock Joint unsupervised learning of deep representations and image
  clusters.
\newblock In \emph{CVPR}, 2016.

\bibitem[Yeh et~al.(2022)Yeh, Hong, Hsu, Liu, Chen, and
  LeCun]{DecoupledContrastive}
Chun-Hsiao Yeh, Cheng-Yao Hong, Yen-Chi Hsu, Tyng-Luh Liu, Yubei Chen, and Yann
  LeCun.
\newblock Decoupled contrastive learning.
\newblock In \emph{ECCV}, 2022.

\bibitem[Zhan et~al.(2020)Zhan, Xie, Liu, Ong, and Loy]{Zhan_2020_CVPR}
Xiaohang Zhan, Jiahao Xie, Ziwei Liu, Yew-Soon Ong, and Chen~Change Loy.
\newblock Online deep clustering for unsupervised representation learning.
\newblock In \emph{CVPR}, 2020.

\bibitem[Zhou et~al.(2017)Zhou, Zhao, Puig, Fidler, Barriuso, and
  Torralba]{Zhou_2017_CVPR}
Bolei Zhou, Hang Zhao, Xavier Puig, Sanja Fidler, Adela Barriuso, and Antonio
  Torralba.
\newblock Scene parsing through ade20k dataset.
\newblock In \emph{CVPR}, 2017.

\bibitem[Zhou et~al.(2022{\natexlab{a}})Zhou, Wei, Wang, Shen, Xie, Yuille, and
  Kong]{zhou2021ibot}
Jinghao Zhou, Chen Wei, Huiyu Wang, Wei Shen, Cihang Xie, Alan Yuille, and Tao
  Kong.
\newblock ibot: Image bert pre-training with online tokenizer.
\newblock In \emph{ICLR}, 2022{\natexlab{a}}.

\bibitem[Zhou et~al.(2022{\natexlab{b}})Zhou, Zhou, Si, Yu, Ng, and
  Yan]{mugs2022SSL}
Pan Zhou, Yichen Zhou, Chenyang Si, Weihao Yu, Teck~Khim Ng, and Shuicheng Yan.
\newblock Mugs: A multi-granular self-supervised learning framework.
\newblock In \emph{arXiv preprint arXiv:2203.14415}, 2022{\natexlab{b}}.

\bibitem[Zhu et~al.(2023)Zhu, Fu, and Wu]{Zhu_2023_ICCV}
Ke Zhu, Minghao Fu, and Jianxin Wu.
\newblock Multi-label self-supervised learning with scene images.
\newblock In \emph{ICCV}, pages 6694--6703, 2023.

\end{thebibliography}
}

\end{document}